\documentclass[journal]{IEEEtran}
\usepackage{cite}

\ifCLASSINFOpdf
   \usepackage[pdftex]{graphicx}

\else
\fi

\usepackage{amsmath}
\usepackage{array}

\ifCLASSOPTIONcompsoc
  \usepackage[caption=false,font=normalsize,labelfont=sf,textfont=sf]{subfig}
\else
  \usepackage[caption=false,font=footnotesize]{subfig}
\fi

\usepackage{bm}
\usepackage{multicol}
\usepackage{multirow}
\usepackage{diagbox}
\usepackage{amsmath,amsfonts,amsthm}
\usepackage{threeparttable}

\usepackage{caption}
\usepackage{cancel,soul}
\usepackage[normalem]{ulem}

\usepackage[colorlinks=true]{hyperref}

\usepackage{color}

\soulregister\cite7
\soulregister\ref7
\soulregister\pageref7

\setulcolor{blue}
\setstcolor{red}

\def\add#1{#1}

\begin{document}
%
\title{Line Flow Based SLAM}
%
%
%

\author{Qiuyuan Wang, Zike Yan, Junqiu Wang, Fei Xue, Wei Ma, Hongbin Zha 
\thanks{This work is supported by the National
Key Research and Development Program of China
(2017YFB1002601) and National Natural Science Foundation
of China (61632003, 61771026).}
\thanks{Qiuyuan Wang, Zike Yan, Fei Xue, and Hongbin Zha are with the Key Laboratory	of Machine Perception (Minister of Education), Peking University, Beijing 100871, China (e-mail: \{wangqiuyuan, zike.yan,  feixue\}@pku.edu.cn, zha@cis.pku.edu.cn).}
\thanks{Junqiu Wang is with AVIC Beijing Changcheng Aeronautical Measurement and Control Technology Research Institute, Beijing 100176, China (e-mail: jerywangjq@foxmail.com).}
\thanks{ Wei Ma is with the Faculty of Information Technology, Beijing University of Technology, Beijing 100124, China (e-mail: mawei@bjut.edu.cn).}
\thanks{Corresponding authors: Wei Ma and Hongbin Zha.}
}

\markboth{IEEE TRANSACTIONS ON ROBOTICS}%
{Wang \MakeLowercase{\textit{et al.}}: LINE FLOW BASED SLAM}

\maketitle


\begin{abstract}

We propose a visual SLAM method by predicting and updating line flows that represent sequential 2D projections of 3D line segments. While feature-based SLAM methods have achieved excellent results, they still face problems in challenging scenes containing occlusions, blurred images, and repetitive textures. To address these problems, we leverage a line flow to encode the coherence of line segment observations of the same 3D line along the temporal dimension, which has been neglected in prior SLAM systems. Thanks to this line flow representation, line segments in a new frame can be predicted according to their corresponding 3D lines and their predecessors along the temporal dimension. We create, update, merge, and discard line flows on-the-fly. We model the proposed line flow based SLAM (LF-SLAM) using a Bayesian network. Extensive experimental results demonstrate that the proposed LF-SLAM method achieves state-of-the-art results due to the utilization of line flows. Specifically, LF-SLAM obtains good localization and mapping results in challenging scenes with occlusions, blurred images, and repetitive textures.

\end{abstract}


%
\IEEEpeerreviewmaketitle

\section{Introduction}\label{sec:introduction}

\IEEEPARstart{S}{imultaneous} localization and mapping (SLAM) continuously estimates camera motions and reconstructs the structure of a scene in an unknown environment. This technique is critical in Unmanned Aerial Vehicle (UAV), autonomous driving, augmented reality, and  robotics applications.

\begin{figure}[!t]
	\centering
	\includegraphics[width=\linewidth]{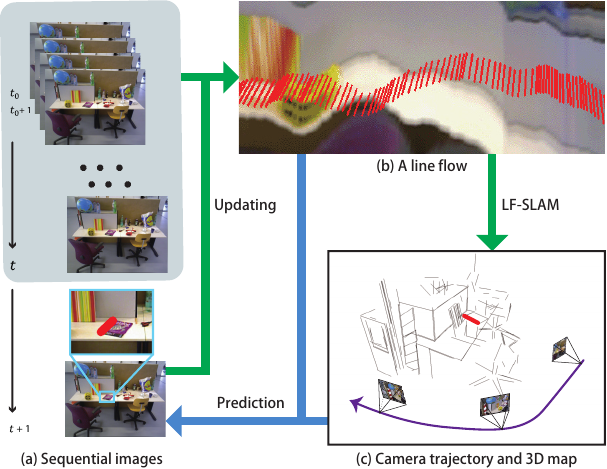}
	\caption{Overview of the LF-SLAM system. The system achieves camera
			localization and 3D mapping (c) with sequentially acquired image frames (a)
			while maintaining a group of line flows (b) on-the-fly. At $t + 1$ frame,
			it predicts a line segment via its corresponding 3D line and preceding
			observations. The prediction is then used to guide the robust detection of the
			line segment in the updating step. Note that (b) is composed of a sequence
			of image slices carrying the 2D observations to visualize a line flow.}
	\label{fig:LineFlowSLAMPipeline}
\end{figure}

SLAM systems can be categorized into direct and indirect (feature based) methods according to their input of optimization~\cite{Engel2018,Engel2014}. Direct methods~\cite{Engel2018,Engel2014,Newcombe2011} leverage raw sensor data to minimize photometric errors. The optimization process also integrates a full photometric calibration, accounting for exposure time, lens vignetting, and nonlinear response functions. Compared to direct methods, indirect methods utilize features, originally considered keypoints~\cite{Mur-Artal2017}, as optimization inputs based on local/global maps, which are thereby more robust to photometric variation and can straightforwardly incorporate loop closure. However, keypoint-based SLAMs often fail in man-made scenarios, which contain many low textured regions and repetitive textures.

Line features widely exist in man-made scenarios. In addition, they are much more informative than points in representing the structure of scenes. Unfortunately, incorporating lines in SLAM is not trivial. Although significant progress has been made in recent years \cite{VonGioi2010,Zhang2013,Almazan2017,Cho2018}, line segment extraction and matching remain challenging. In particular, the endpoints of detected line segments are usually unstable across sequential frames. In addition, a threshold for line segment extraction is not easily determined for reducing false positives while increasing recall. An improper threshold, and other traits, such as occlusions or appearance ambiguities, can break a line into several smaller lines. Other issues can also arise when lines are broken, thereby deteriorating SLAM performance in various ways, such as a) unstable 2D endpoints from broken segments causing drift in endpoint triangulation and b) smaller segments cluttering the reconstructed map with redundant 3D line segments. In addition, descriptor-based matching methods~\cite{Zhang2013, Pumarola2017, Zhao2018_ECCV} are prone to ambiguous matches when the scene is rich in repetitively textured surfaces.

To address the above challenges, we propose line flows to exploit both the spatial and temporal coherence of line segment motions in 2D and 3D domains; that is a line segment extracted from a previous frame provides valuable priors for its extraction in the current frame, ensuring a highly repeatable extraction process. Spatial and temporal coherence is inherent in SLAM but not fully exploited by existing systems. Some SLAM systems simply leverage the small motion assumption as smoothness priors to reduce time costs. For example, ORB-SLAM2~\cite{Mur-Artal2017} and DSO~\cite{Engel2018} build grids to decrease candidates in the feature matching step. Additionally, based on the small motion assumption, some systems take the camera pose of a previous frame as that of the current frame, followed by Gaussian-Newton or Levenberg-Marquardt algorithms to refine the pose. These strategies work in practice because the temporal smoothness prior is valid in most modern cameras. In this work, we exploit the spatial and temporal coherence of motions more thoroughly via the proposed line flow features.

We then develop a line flow based SLAM (LF-SLAM) by formulating the line flows using a Bayesian network. As illustrated in Fig.~\ref{fig:LineFlowSLAMPipeline}, a line segment in a newly acquired frame is first predicted by the projection of its corresponding 3D line, reconstructed before this moment, and its predecessor in the previous frame. Guided by this spatial projection and temporal prediction, line extraction in the current frame is more efficient in computation, and stable in direction and endpoint positions, even when facing temporal occlusions, repetitive textures and blurred images. The consecutive observations of the 3D line segment form a line flow.

The contributions of this work are twofold:
\begin{itemize}
	\item[-] We propose line flows to model the spatial-temporal coherence of line segments. The line flows are maintained in a predicting-updating fashion. Due to the coherence constraints, the extracted line observations are stable across sequential frames even when facing temporal occlusions, blurred images, and repetitive textures;
	
	\item[-] We develop a monocular SLAM system based on the line flow representation using a Bayesian network. We create, insert, merge, and discard line segments at each frame based on the line flow representation, to bridge the different stages of a SLAM system, including feature extraction, feature matching, triangulation, and optimization. Compared to the other SLAM systems with lines~\cite{Gomez-Ojeda2017,Pumarola2017}, our LF-SLAM fully explores the temporal-spatial coherence of line observations via line flows and is thus more efficient in computation and more accurate in camera localization. In the meantime, our system generates visually appealing 3D maps on-the-fly.
\end{itemize}

We organize this work as follows. We introduce \add{related} work in Section~\ref{Related Work}. The line flow representation and the SLAM system pipeline are described in Section~\ref{Sec:LineFlowRepresentation}. Then, the details of line flow tracking and mapping are provided in Section~\ref{Sec:LineFLowTracking} and Section~\ref{Sec:LineFlowMapping}, respectively.
Extensive experiments, including comparisons with state-of-the-art SLAM approaches are  reviewed in Section~\ref{Sec:Experiment}. Finally, we conclude this work in Section~\ref{Sec:Conclusion}.

\section{Related Work}\label{Related Work}
The proposed LF-SLAM requires various sub-tasks, namely line segment detection and matching, 3D line reconstruction, and visual SLAM. There are a large number of literatures on each subtask. We briefly review the most relevant works.

\subsection{Line Segment Detection and Matching} Detecting and matching line segments from images is a classical problem, which is essential for diversified 3D vision tasks, e.g., line reconstruction and SLAM. Hough transform~\cite{Ballard1981,Almazan2017} is widely used for line detection by voting  in transformed spaces. However, it is computationally expensive, hence, not suitable for real-time applications. Cho \textit{et al.}~\cite{Cho2018} gives a comprehensive analysis of line segments and presents a novel linelet representation for line segment detection. Although effective, its high computational cost restricts its practical usage. Recently, learning-based detection methods~\cite{Xue_2019_CVPR,Zhang_2019_CVPR} have emerged and attracted much attention.
These methods generally rely on GPUs for training and \add{inference}.
In addition, \add{they cannot guarantee the stableness of} line detections across sequential frames. Line Segment Detector (LSD)~\cite{VonGioi2010} and its improved versions are widely used in SLAM~\cite{Pumarola2017} and SfM systems~\cite{Salaun2016}. These algorithms locates line segments by aggregating adjacent pixels with similar gradient orientations~\cite{VonGioi2010,Akinlar2011,Salaun2016}. However, the involved gradient pseudo-ordering and region growing computations are still complex and the greedy growing often leads to broken lines. The proposed LF-SLAM also adopts LSD. Differently, we consider the spatial-temporal coherence of line segments during detection in image sequences to improve the computational efficiency and stability of line segments across frames.

Many methods have been proposed for line segment matching. For example, MSLD \cite{Wang2009} calculates the mean and standard deviation statistics of each line for matching. The approaches in~\cite{Fan2012, Micusik2017} match lines according to their neighboring feature descriptors. LBD \cite{Zhang2013}, which is widely adopted in SLAMs~\cite{Pumarola2017,Gomez-Ojeda2017,Zhao2018_ECCV}, builds a relational graph and leverages local appearance and geometry relationship for line matching.
Instead of using 2D spatial context for matching, the proposed LF-SLAM fully exploits the spatial-temporal coherence by line flows to guide efficient and effective line segment matching.

\subsection{Visual SLAM with Lines}

LineSLAM~\cite{Perdices2014} takes lines as basic features and \add{tracks} them via unscented Kalman filter (UKF). It shows good performance in simulated environments. StructSLAM~\cite{Zhou2015} uses building structure lines, which encode global orientation, to eliminate accumulated errors and drift in Manhattan-like environments. Li \textit{et al.}~\cite{Li2018} also leverage structural line features to compute camera poses. Sol{\`{a}} \textit{et al}.~\cite{Sola2012} conduct a comprehensive study to better understand the impact of different point and line parametrization on SLAM based on Extended Kalman Filter (EKF). However, the complexity of the EKF algorithm depends on the square of the number of the landmarks, which leads to unacceptable computational cost for real-time applications.

Dong \textit{et al.}~\cite{Ruifang2018} propose a monocular SLAM method using point and line features for graph optimization, which achieves higher accuracy than the EKF-based SLAM method~\cite{Sola2012}. In a similar way, PL-SLAM Mono~\cite{Pumarola2017} performs beyond many prior and contemporary methods. PL-SLAM Mono extracts line segments and their descriptors for matching. The complexity of line segment detection and matching is relatively high since all the pixels in an image need to be visited for LBD descriptors. As to bundle adjustment, 3D lines from local map are projected onto images to search for correspondences. Bundle adjustment utilizes points and lines together in point-line re-projection errors. Loop closure only uses points in PL-SLAM Mono~\cite{Pumarola2017}, since matching lines across the whole map is too computationally expensive. In addition, the detected lines are unstable across frames. Zhao \textit{et al}.~\cite{Zhao2018_ECCV} cuts unreliable line parts to eliminate their influences on localization. However, while boosting the performances in localization, their method neglects the roles of the lines in the 3D structure representation of physical scenes. Our method achieves efficient SLAM with high localization accuracy and concise 3D maps due to the unified framework which fully exploits the spatial-temporal cues induced by the proposed line flow representation. In addition, once we detect a loop closure, we can optimize global parameters based on the long-term stable 2D line segments recorded in the line flows.

\begin{figure}[!t]
	\centering
	\includegraphics[width=\linewidth]{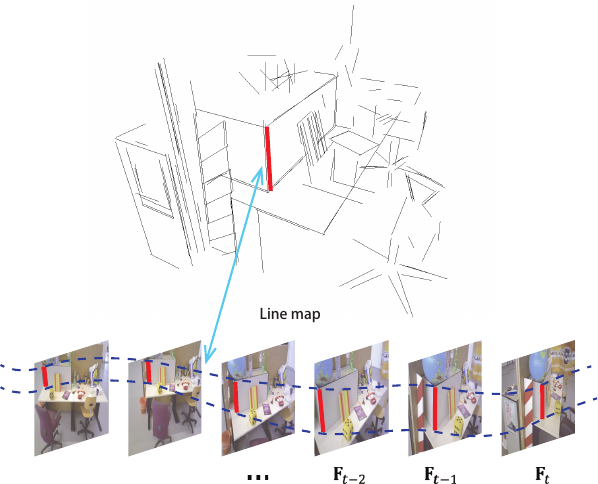}
	\caption{A line flow in sequential images. All the line segments in the line flow correspond to the same 3D line.}
	\label{fig:LineFlowConception}
\end{figure}

\subsection{3D Line Reconstruction} 3D line reconstruction has been studied for decades. An early attempt~\cite{Ayache1987} uses a graph-based description of line segments for 3D line reconstruction from two stereo images. Bartoli \textit{et al.}~\cite{Bartoli2005} present a line-based SfM pipeline. However, the strong constraint within the trifocal tensor restricts the reconstructed lines. Jain \textit{et al.}~\cite{Jain2010} impose global topological constraint by using neighboring connections between line segments for outlier rejection, but it requires ground truth of camera poses. Recently, He \textit{et al.}~\cite{He2017} leverage the collinear property from extracted 2D line segment cues to fit 3D lines. Hofer \textit{et al.}~\cite{Hofer2017} formulate line reconstruction procedure as a graph-clustering problem. These methods achieve promising results. Unfortunately, they all assume a known camera pose which is not suitable for SLAM.

All the aforementioned methods require the entire image sequence input in a batch mode, and therefore have to run off-line. Some incremental 3D line reconstruction methods are proposed for online applications. For example, Zhang \textit{et al.}~\cite{Zhang2014} complement the Pl{\"{u}}cker coordinate with the Cayley representation, but fail to handle the non-trivial line matching and translation motion drift issues. The approach in~\cite{Micusik2017} tries to deal with unstable endpoints, and gains efficiency by decoupling translation and rotation. However, it relies on a strong assumption that two parallel lines are orthogonal to a third one, which restricts its practical usage.

\begin{figure}[!bp]
	\centering
	\includegraphics[width=\linewidth]{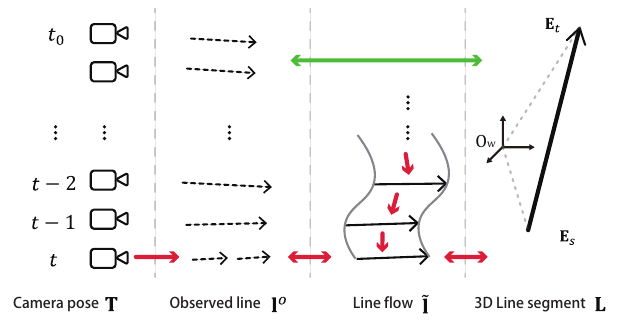}
	\caption{
	Coherence within a line flow. 
	Traditional methods only establish correspondences between observed line segments $\mathbf{l}^o$ and 3D lines $\mathbf{L}$ at each view (green arrows). In contrast, the line flow $\tilde{\mathbf{l}}$ establishes additional temporal correlation between line segments at different views that correspond to the same 3D line segment (red arrows). 
	As a result, line parameters are constrained with priors. Broken line segments can be merged based on the priors.
	 }
	\label{fig:LineFlowModel}
\end{figure}

\section{Line Flow \& Its Formulation in SLAM} \label{Sec:LineFlowRepresentation}
We introduce the proposed line flow representation in Section \ref{SubSec:LineFlowRepresentation} and its formulation in SLAM in Section~\ref{SubSec:LineFlowBasedSLAM}.

\subsection{Line Flow Representation} \label{SubSec:LineFlowRepresentation}
As illustrated in Fig.~\ref{fig:LineFlowConception}, a line flow $\tilde{\mathbf{l}}$ is defined as the sequence of 2D line segments  corresponding to the same 3D line $\mathbf{L}$ and is maintained from the first observation at time $t_0$ to the current frame at $t$:
\begin{equation}
\tilde{\mathbf{l}} = \{ \mathbf{l}_{t_0}, \mathbf{l}_{t_0+1}, \cdots, \mathbf{l}_t \}.
\end{equation}

We parameterize a 2D line segment $\mathbf{l}$ by $(\theta,l,x,y)$, including the line orientation $\theta$, length $l$, and middle point position $(x,y)$.
 We adopt the orientation representation in ~\cite{VonGioi2010}.
Following~\cite{Hofer2017} and \cite{Zhang2015},  an oriented 3D line segment $\mathbf{L}$ is parameterized using a 3D infinite line representation $(\begin{matrix}\mathbf{n}^T,\mathbf{v}^T\end{matrix})^T$ \cite{Hartley2004} in the Pl{\"{u}}cker coordinates with initial and terminal endpoints $\mathbf{E}_s$, and $\mathbf{E}_t$.

Following the definition, we summarize 3 properties:
\begin{enumerate}
	\item \textbf{The relationship between a 2D line segment and its corresponding 3D line segment.} Each 2D line segment $\mathbf{l}_{t_i}$ in the line flow is collinear with the 2D projection of the 3D line segment. Note that in practice, the 2D projection of a 3D line segment can be different from the observed 2D line segments in incoming images due to occlusions, appearance ambiguities, etc. Based on this property, we design the back-end optimization module of our SLAM system (in Section~\ref{SubSec:LineFlowBasedSLAM} \& Section~\ref{Sec:LineFlowMapping}) to solve the above problem.
	
	\item \textbf{The relationships of 2D line segments in a line flow.} As observations of the same 3D line in consecutive frames, 2D line segments are formed due to camera motion. The inherent spatial-temporal coherence constraints among the 2D line segments provide helpful information for line flow extraction. We design a line flow prediction module by leveraging the constraints (in Section~\ref{Prediction}).
	
	\item \textbf{Line segment properties.} Each line in a line flow inherits the properties of line segments; a) each has a unique direction, b) each line segment is observed as a set of pixels with similar gradient orientations, and c) the gradient orientations are perpendicular to the line segment. Based on these properties, we can merge nearby subsegments with similar line directions (in Eq.~\eqref{eq:fusion}) and guide line extraction by grouping similar gradient orientations (in Section~\ref{2D Correction}).
\end{enumerate}

We show how to build line flows using the spatial-temporal coherence in Fig.~\ref{fig:LineFlowModel}. It can be seen that the coherence of line segments in consecutive frames can help solve the incorrect \add{observations} in single frames \add{that may be} caused by occlusions, image blur, and noise.

\begin{figure}[t]
	\centering
	\includegraphics[width=\linewidth]{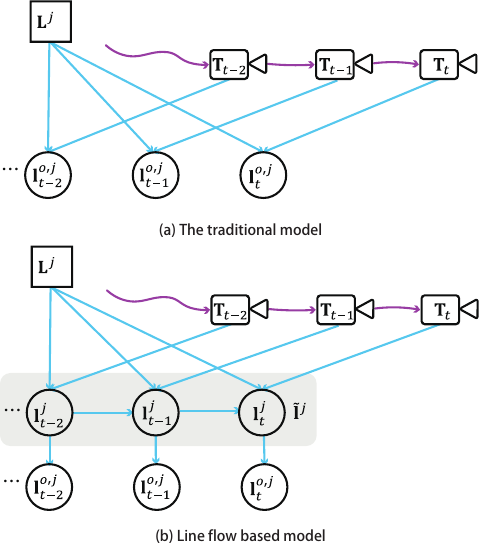}
	\caption{Compared to the traditional SLAM model (a), our line flow based model (b) establishes correlations for \add{temporally adjacent line segment observations and camera poses. $\mathbf{L}^{j}$ denotes the $j$-th 3D Line. Its line flow is $\mathbf{\tilde{l}}^{j}=\{\mathbf{l}^{j}\}$. $\mathbf{l}^{o,j}$ with a subscript, e.g., $t$, denotes the observation of $\mathbf{L}^{j}$ at frame $t$. $\mathbf{T}_t$ represents the camera pose at frame $t$.}}
	\label{fig:LineFlowGraphModel}
\end{figure}

Finding reliable line segment correspondences is a prerequisite for successful triangulation. Existing line-based SLAM methods only consider the correspondences of line segments in two frames during line detection and matching~\cite{Gomez-Ojeda2017,Pumarola2017,Zhang2015}. Unfortunately, finding reliable correspondences in this way is very difficult because a) partial occlusions often split a line into fragments, b) image noise might lead to unstable line endpoints, c) visual artefacts generally bring about false-positive line segments, and  d) a scene with repetitive textures may cause severe matching ambiguities.

The proposed line flow exploits spatial-temporal consistency to establish correspondences between sequential line segments for 2D line prediction, extraction, updating and matching. Therefore, this technique has several advantages:
\begin{figure*}[t]
	\centering
		\includegraphics[width=\linewidth]{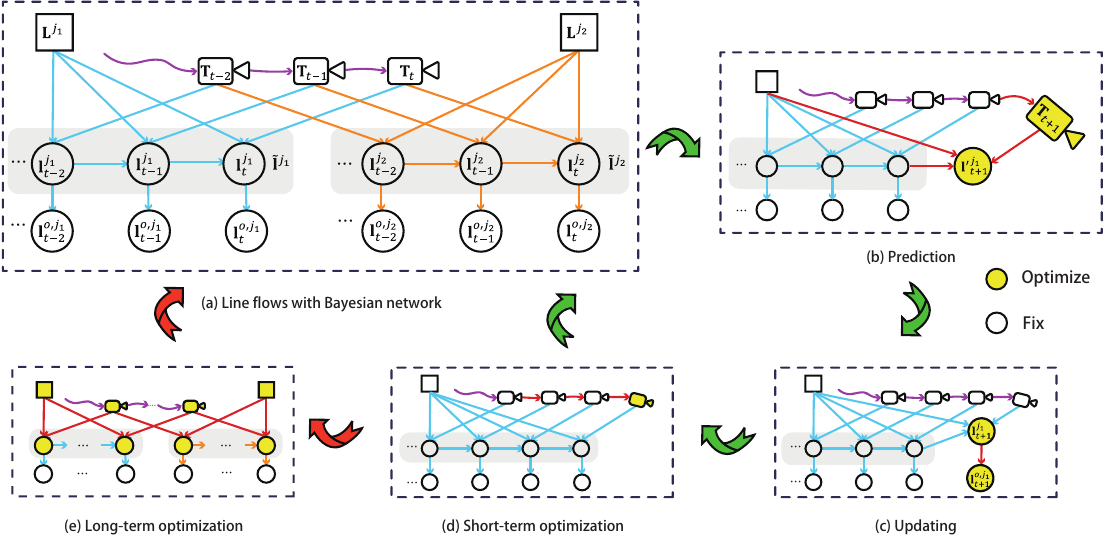}
		\caption{We model the SLAM problem using a Bayesian network (a), which consists of line segment prediction (b), line flow updating (c), short term optimization (d), and long term optimization (e). The yellow nodes denote variables to be updated, and \add{the} white nodes are fixed. The processes indicated by the green arrows are processed frame-by-frame, while those indicated by the red arrows are performed only on key frames.
		}
	\label{fig:PipeLine}
\end{figure*}

\begin{enumerate}
\item We do not need an explicit line segment descriptor
 because we predict and update line segments in a line flow based on spatial-temporal coherence;
\item The constraints in a line flow are utilized for reliable and efficient 2D line segment extraction. Extracting line segments and finding their correspondences in this way are robust against occlusions and outliers;
\item The line flows help reject false positive observations. For example, the sides of a cylinder, not coherent in consecutive frames, are discarded.
\end{enumerate}

\subsection{Line flow based SLAM} \label{SubSec:LineFlowBasedSLAM}

As illustrated in Fig. \ref{fig:LineFlowGraphModel}, we construct a graph model for line flows. Each line flow $\tilde{\mathbf{l}}^j$ corresponds to a 3D line $\mathbf{L}^j$, and the correlation is constrained by camera poses $\mathbf{T} \in \mathrm{SE(3)}$ and the observations $\mathbf{l}_t^o$.
At the very beginning of a SLAM process, we take the camera coordinate of the first frame as the global coordinate. The relative transform from the global coordinate to the $t$-th frame is denoted by $\mathbf{T}_{t} \in \mathrm{SE}(3)$.

As illustrated in Fig.~\ref{fig:PipeLine}~(a),
the probabilistic relationships between the variables and the observations can be modelled by a Bayesian network. The joint probability can be described as:
\begin{equation}
	\begin{aligned}
	\mathrm{P}\big( S
	\big)
	\propto \mathrm{P}\big(\mathbf{T}_0 \big)
			\prod_{t,j,k
	} &\mathrm{P}\big( \mathbf{l}^{o,j}_t \vert \mathbf{T}_t,\mathbf{L}^j,\mathbf{l}^j_{t-1}  \big) \\
	&\mathrm{P}\big(\mathbf{p}^{o,k}_t \vert \mathbf{T}_t, \mathbf{P}^k\big)
	\mathrm{P}\big(\mathbf{T}_t \vert \mathbf{T}_{0 ... t-1}\big)
	,
	\end{aligned}
	\label{ProblemFormulationFormat}
\end{equation}
where $S$ denotes the entire set of all unknown variables including camera poses $\mathbf{\{T\}}$, 3D line segments $\{\mathbf{L}\}$, 2D line flows $\{\tilde{\mathbf{l}}\}$, and 2D line observations $\{\mathbf{l}^o\}$;
$\mathrm{P}\big(\mathbf{T}_0 \big)$ is the camera pose prior;
$\mathrm{P}\big( \mathbf{l}^{o,j}_t \vert \mathbf{T}_t,\mathbf{L}^j,\mathbf{l}^j_{t-1} \big)$ is the line measurement model; $\mathrm{P}\big(\mathbf{p}^{o,k}_t \vert \mathbf{T}_t, \mathbf{P}^k\big)$ is the point measurement model; and $\mathrm{P}\big(\mathbf{T}_t \vert \mathbf{T}_{0 ... t-1}\big)$ is the camera motion model.
The probabilistic formulation in Eq.~\eqref{ProblemFormulationFormat} can be transformed to an energy function and can be solved efficiently with a nonlinear optimization method~\cite{Kaess2008,Thrun_2005,Zhang2015}.

The proposed pipeline is illustrated in Fig.~\ref{fig:PipeLine} in an incremental fashion.
We design a front-end (green arrows) running incrementally at frame-rate for optimizing line flows and camera poses and a back-end (red arrows) running only for key-frames to optimize the 3D line map and camera poses.
Fig.~\ref{fig:PipeLine}~(a) is the graph state at time $t$.
We predict the 2D projection of the 3D line segment corresponding to a line flow in a new frame. Then, we update the line flow according to the constraints and the information in a new observation.
We optimize the camera pose $\mathbf{T}_{t+1}$ by fixing 2D features and 3D landmarks in the short-term optimization. We also perform long-term optimization for the 3D line map in the back-end.

\add{we adopt ORB-SLAM in~\cite{Mur-Artal2017} as a base
framework in which we develop alongside its 3D point map, our line map.
We also extend ORB-SLAM's pose estimation and mapping processes to make
use of both points and lines as described in sec V-A and sec V-B,
respectively.}
We extract ORB features from each image.
Features are separated into cells equally dividing the images.
We calculate ORB descriptors for point matching.
We estimate a robust fundamental matrix or homography matrix with RANSAC~\cite{Hartley2004}. The 3D points are triangulated to check rotation and translation decomposed from the fundamental matrix or homography matrix. Initialization is achieved when the inlier number of 3D points is greater than a set threshold. We initialize the succeeded frames as keyframes.
For the following frames, a two-step strategy is performed to enhance the robustness and efficiency of our system.
First, we leverage the predicted camera poses $\mathbf{T}_{t+1}$ as coarse poses to search for 2D candidates near the 2D projected point of each 3D point. A grid strategy is applied to accelerate this procedure by dividing images into cells and keypoints are collected from their the corresponding cells. We optimize the predicted camera poses $\mathbf{T}_{t+1}$ with the point correspondences.
Second, we search for more correspondences by projecting 3D points from local maps.
These 2D-3D correspondences are then applied in short-term optimization.
In long-term optimization, 3D points are
triangulated when a new keyframe is created. We use geometric checking to find duplications.
Since the procedures for line and point features are independent, we manage them in parallel.

\section{Line Flow Tracking}\label{Sec:LineFLowTracking}

\begin{figure*}[htbp]
	\centering
	\subfloat[RGB image]{
		\begin{minipage}[c]{.23\linewidth}
			\label{subfig:rgb}
			\centering
			\includegraphics[width=\linewidth]{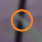}
		\end{minipage}
		\label{subprediction_a}
	}
	\hfil
	\subfloat[Level-line field]{
		\begin{minipage}[c]{.23\linewidth}
			\label{subfig:rectangle}
			\centering
			\includegraphics[width=\linewidth]{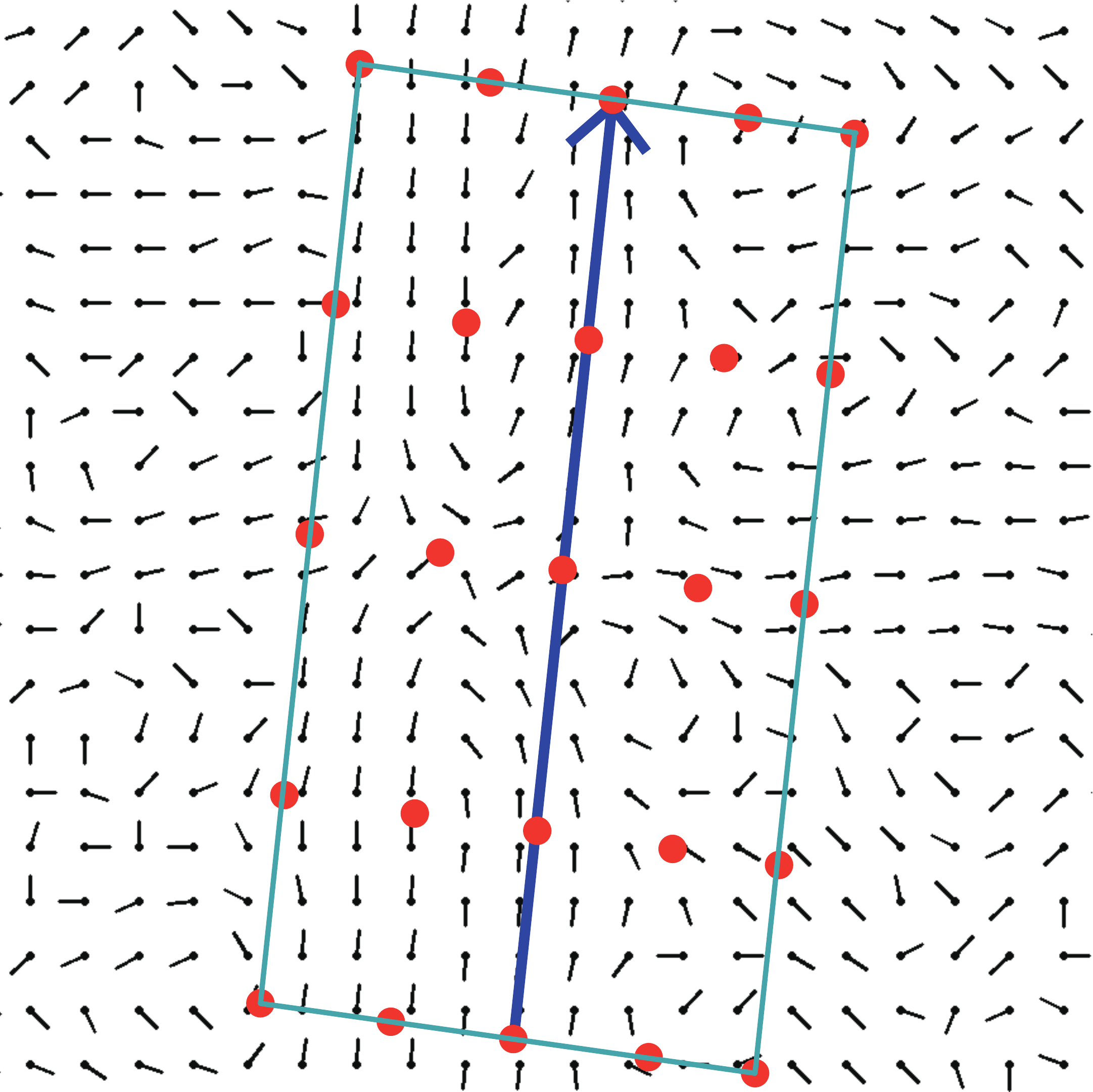}
		\end{minipage}
		\label{subprediction_b}
	}
	\hfil
	\subfloat[Line support regions]{
	\begin{minipage}[c]{.23\linewidth}
		\label{subfig:LSR}
		\centering
		\includegraphics[width=\linewidth]{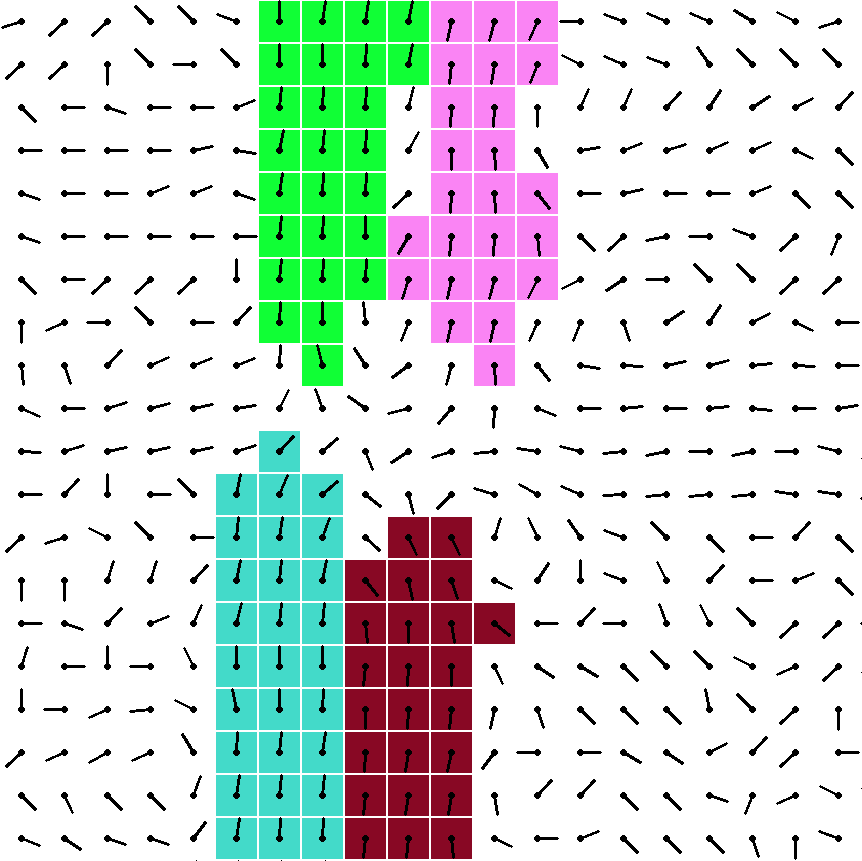}
	\end{minipage}
		\label{subprediction_c}
	}
	\hfil
	\subfloat[Line fusion]{
	\begin{minipage}[c]{.23\linewidth}
		\label{subfig:fusion}
		\centering
		\includegraphics[width=\linewidth]{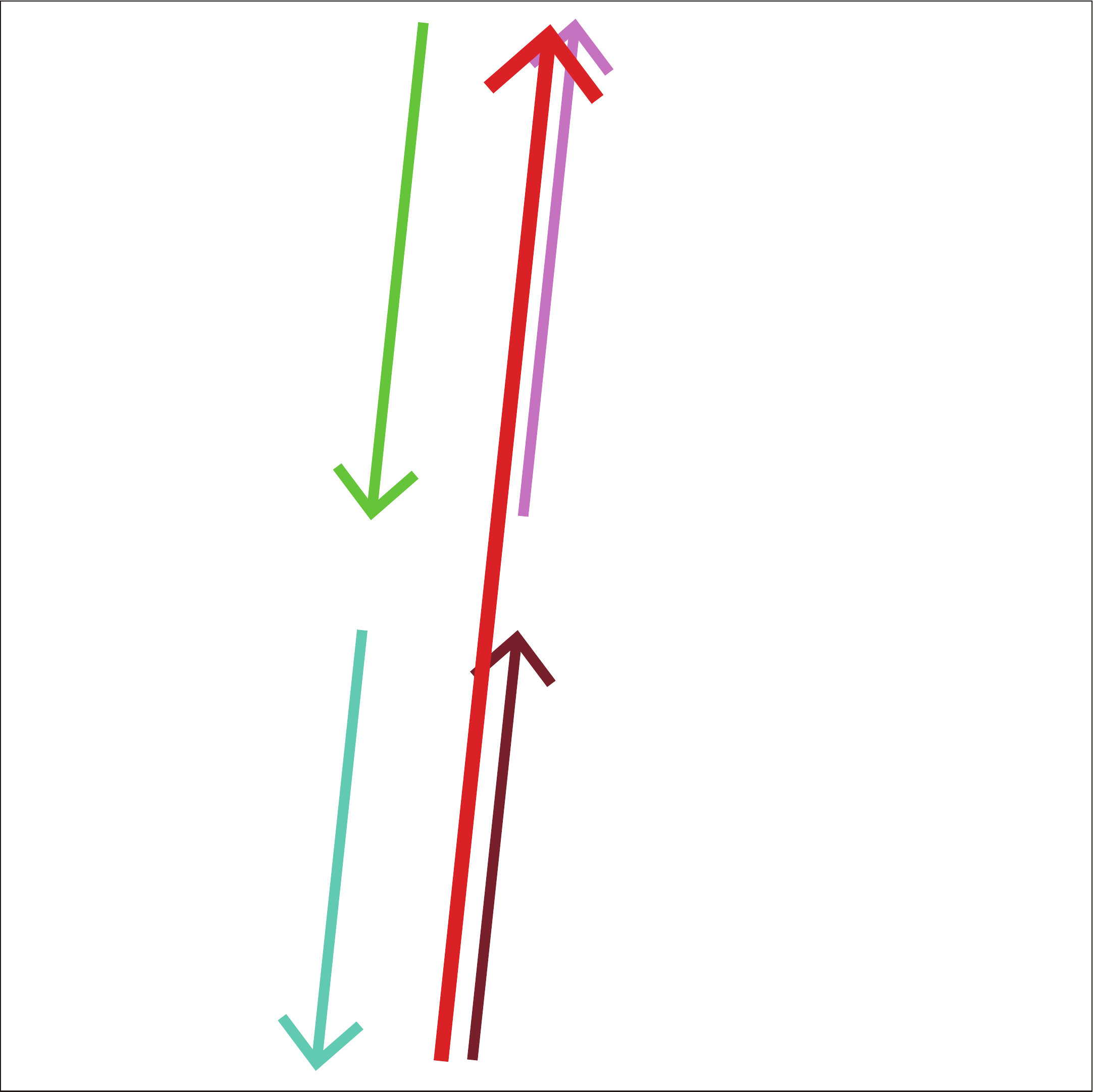}
	\end{minipage}
		\label{subprediction_d}
	}
	\caption{Guided line extraction in  the circumstance of occlusions (a). A few seeds (red) are uniformly sampled (b) within the rectangular area around the predicted line segment (dark blue), where level-line field~\cite{GromponevonGioi2012} is illustrated as small black vectors. Four line support regions are found using the seeds (c), and four line segment candidates are subsequently found. Line segment that is the most collinear to the predicted line is chosen as the best candidate. Then, we fuse it with other collinear candidates. In this way, we can extract the complete line segment even when occlusion occurs (d), and we can distinguish line segments on both sides according to our similarity criteria. The line segment on the left side will be extracted with the corresponding predicted line.
	}
	\label{fig:LineFlowPrediction}
\end{figure*}

In this section, we introduce how to track line flows by performing prediction and updating in the front-end. The main task of the \add{line} tracking module is to maintain line flows when new frames are captured. To achieve this, we first predict each line segment by finding the correspondences of line segments in 2D and 3D domains based on camera motion and 3D projection. A line flow is then updated based on a new observation. Unlike other line-based SLAM systems, we can find reliable correspondences using the spatial-temporal constraints and line geometric properties. 

\subsection{Prediction} \label{Prediction}
We infer a new line segment based on motion coherence. The first derivatives of the 6-DoF camera motion, denoted as $\mathbf{m}_{\mathbf{T}}$, and the differences between parameters of consecutive 2D line observations, denoted as $\mathbf{m}_{\mathbf{l}}$, are assumed to be constant within a temporal sliding window. Starting from the $(t-t_w)$-th frame, we have
\begin{equation}
	\begin{aligned}
	\mathbf{T}_{t} &= \mathbf{T}_{t-1} \mathbf{m}_{\mathbf{T}}, \\
	\mathbf{T}_{t-1} &= \mathbf{T}_{t-2} \mathbf{m}_{\mathbf{T}},\\
	\cdots \\
	\mathbf{T}_{t-t_w+1} &= \mathbf{T}_{t-t_w} \mathbf{m}_{\mathbf{T}}.
	\end{aligned}
\end{equation}
Similar to pose motion,  we have
\begin{equation}
\begin{aligned}
\mathbf{l}_{t}^j &= \mathbf{l}_{t-1}^j + \mathbf{m}_{\mathbf{l}}^j, \\
\mathbf{l}_{t-1}^j &= \mathbf{l}_{t-2}^j + \mathbf{m}_{\mathbf{l}}^j,\\
\cdots  \\
\mathbf{l}_{t-t_w+1}^j &= \mathbf{l}_{t-t_w}^j + \mathbf{m}_{\mathbf{l}}^j,
\end{aligned}
\end{equation}
We calculate these two motions by solving a least-square problem~\cite{Thrun_2005}.

\noindent\textbf{Line segment prediction  based on 2D coherence.}
2D prediction is achieved by modelling the 2D motion $\mathbf{m}_\mathbf{l}^j$ of each line flow $\tilde{\mathbf{l}}^j$. 
With the aforementioned constant velocity assumption, the 2D motion model of each line flow can be expressed with previous estimation results, and therefore the line segment $\mathbf{g}_{t+1}^j$ predicted by 2D coherence is given by:
\begin{equation}
\mathbf{g}_{t+1}^j = \mathbf{l}_{t}^j+\mathbf{m}^j_{\mathbf{l}}.
\end{equation}

\noindent\textbf{Line segment prediction based on 3D coherence.} 3D prediction is achieved by projecting two endpoints of each 3D line segment $(\mathbf{E}^j_\mathbf{s},\mathbf{E}^j_\mathbf{t})$ into the current frame with a predicted camera pose $\mathbf{T}_{t+1}= [ \mathbf{R}_{t+1} \vert \mathbf{t}_{t+1}]$,  which is calculated in a similar way as the 2D motion model. Then, the line segment $\mathbf{h}_{t+1}^j$ predicted by using 3D coherence is calculated by:
\begin{equation}
\begin{aligned}
\mathbf{e}_\mathbf{s}^j &=\lfloor \mathbf{K}_p ( \mathbf{R}_{t+1} \mathbf{E}_\mathbf{s}^j + \mathbf{t}_{t+1} ) \rfloor,\\
\mathbf{e}_\mathbf{t}^j &=\lfloor \mathbf{K}_p ( \mathbf{R}_{t+1} \mathbf{E}_\mathbf{t}^j + \mathbf{t}_{t+1} ) \rfloor,\\
\mathbf{h}_{t+1}^j &\leftarrow (\mathbf{e}^j_\mathbf{s},\mathbf{e}^j_\mathbf{t} ),
\end{aligned}
\end{equation}
where $\mathbf{K}_p$ is the intrinsic matrix for point features. $\lfloor \cdot \rfloor$ represents the transformation from the homogeneous coordinate to the 2D coordinate. $\mathbf{e}_\mathbf{s}^j$ and $\mathbf{e}_\mathbf{t}^j$ are two projected endpoints.

\noindent\textbf{Line segment prediction based on both 2D and 3D coherence.}
Line segment prediction using 2D coherence is performed for all the frames. This prediction is useful when 3D line segments are not stable (e.g., during the initialization of a line flow). However, 2D motion might be unreliable due to large displacement. Fortunately, line segment prediction using 3D coherency is helpful for improving the accuracy since camera trajectories are usually smooth.
In general, a long line segment potentially indicates accurate line directions.  Therefore, we integrate the two types of prediction results via a length-based weighting operation:
\begin{equation}
\begin{aligned}
\mathbf{l'}_{t+1}^j
= \frac{l_\mathbf{h}}{l_\mathbf{h} +l_\mathbf{g}} \mathbf{h}^j_{t+1} +
\frac{l_\mathbf{g}}{l_\mathbf{h} +l_\mathbf{g}} \mathbf{g}^j_{t+1},
\end{aligned}
\label{eq:fusion}
\end{equation}
where $l_\mathbf{h}$ and $l_\mathbf{g}$ are the lengths of $\mathbf{h}^j_{t+1}$ and $\mathbf{g}^j_{t+1}$, respectively.
A long line segment has more influence on the line fusion results, which means that the fused line angle leans towards the longer line segment.

When no prior motion is available for a line segment, we directly use it as a predicted line segment. We perform KLT~\cite{JianboShi1994} for each seed generated by the predicted line segment. Then, the corresponding seeds are adopted to guide line extraction.

The purpose of line segment prediction is to accelerate the line extraction and matching processes and to help fuse broken line segment detections. In addition, when we cannot locate observations due to image blur or false positive detection, we directly use the predicted line segments to maintain the continuity of their corresponding line flow. With observations, the predicted line segments are refined before being integrated into line flows.

\subsection{Updating} \label{2D Correction}
We update a line flow with its new prediction and observations in a new frame. The updating has 3 steps: guided line extraction, line flow creation, and line flow management.

\noindent\textbf{Guided line extraction.}
The extraction process is illustrated in Fig.~\ref{fig:LineFlowPrediction}. We highlight the major difference between our guided line extraction method and other methods~\cite{Pumarola2017,Zhao2018_ECCV} that extract line segments independently on each frame using LSD~\cite{VonGioi2010}.

Here, we give a brief description of LSD~\cite{VonGioi2010}. First, the gradient of each pixel is calculated. The gradients in a region form a level-line field~\cite{VonGioi2010}. All the pixels are sorted according to gradient magnitudes. The pixels with high gradient magnitudes tend to form line segments. Starting from the highest gradient pixel as a seed, LSD performs region growing according to the expansion criterion to form line support regions. The expansion criterion is that a pixel must not be visited before and the orientation difference between the angles of the pixels and the angles of the region should be lower than a threshold. The rectangular approximation is applied to form a line segment. Finally, the number of false alarms (NFA) is calculated and refinement is performed for better line segments.

We extract line segments by using the coherence in the spatial-temporal domain to guide the LSD. With line segment prediction, we do not have to detect line segments in a new frame from scratch. Only the nearby regions of the predicted line segments are taken into consideration. This strategy makes our methods efficient because these edge pixels are just a few percent of the images.

To form a line support region, we do not need to sort gradient magnitudes. We adopt a rectangular search region expanded from the predicted line segment. Seeds of line segments are sparsely sampled near the predicted line segments. For each seed, we  grow the region to find the corresponding line segment regions. Then, we extract line segment candidates from each line support region by approximating rectangles according to gradient orientations.
Several line segment candidates are calculated based on the rectangles. We select the one that has the best collinearity with the predicted line segment as the best candidate.
Other candidates can be fused with the best candidate when their angular differences are less than a threshold by performing the fusion operation in Eq.~\ref{eq:fusion}.
With the guidance provided by the prediction, our line extraction rejects many  false-positives and obtains more reliable line segments. Moreover, it can fuse line segments that are broken due to partial occlusions.

\noindent\textbf{Line flow creation.} A line flow is created when we bootstrap the
\add{LF-SLAM} system or when a new line segment is observed. We run the LSD algorithm~\cite{VonGioi2010} for the line extraction of every $N_{cl}$ frame over the remaining unvisited pixels. An experiment (in Section~\ref{Sec:Experiment}) is performed to study the effect of $N_{cl}$. We create a line flow when a new observed line segment comes in.

\noindent\textbf{Line flow management.} We need to carefully maintain stable coherence and enhance robustness. When the 2D line segments corresponding to a line flow are temporarily unobservable in  a few frames, we set a reserving period $\alpha$ before terminating this line flow. Within the reserving period, the line flow is still kept alive with the predicted line segments as the observations.

One 3D line might be split into multiple line flows due to occlusions. We check all line flows and fuse those with more than half of their line segments as collinear and  whose recent line segments overlap with a certain number of pixels (set to 5 pixels).
If two line flows correspond to the same 3D line, we merge them using the merging operation (Eq.~\ref{eq:fusion}).
To accelerate this step, we utilize the collinearity of two line segments in the  current frame. The angle-grid based method is adopted. We set $30^\circ$ as the grid size and put line segments into the corresponding grids between $0^\circ$ and $360^\circ$. Note that the number of line
flows is not very large; therefore, the line flow merging has a very slight negative effect on real-time performance.

Although the orientation of a line segment is relatively stable, its length can vary significantly across different viewpoints. We refine the length \add{by referring to} the historical lengths:
\begin{equation}
	l_{t+1}^j=\max(\frac{1}{\beta} \overline{l}_{t+1}^j,\min (\beta \overline{l}_{t+1}^j,l^{o,j}_{t+1})),
\end{equation}
where $l^{o,j}_{t+1}$ denotes the length of an observed line segment; $\overline{l}_{t+1}^j$ denotes
the mean length of the latest line segments stored in the
corresponding line flow. \add{$l_{t+1}^j$ is controlled by parameter $\beta$, to avoid abnormal length due to wrong observation $l^{o,j}_{t+1}$: if $l^{o,j}_{t+1}$ is much shorter or longer than the mean length, $l_{t+1}^j$ would be restricted by $\beta \overline{l}_{t+1}^j$  or  $\frac{1}{\beta} \overline{l}_{t+1}^j$. $\beta$ is experimentally set to 0.8. A smaller or larger $\beta$ will weaken the motion constraint in a line flow.
}

\section{Line flow based SLAM System}\label{Sec:LineFlowMapping}
SLAM is carried out using line flows with spatial-temporal coherence. The camera pose is estimated for each frame in the short-term optimization, while the 3D map and poses are jointly refined with each key frame through a long-term optimization in a parallel back-end thread. \add{3D line segments, as the 3D counterpart of line flows, will help filter outlier observations in subsequent frames as explained in Section~\ref{Sec:LineFLowTracking} and merge temporally broken line flows as described in Section~\ref{Long Term Optimization}}.

\begin{figure}[t]
	\centering
	\subfloat[Line re-projection error]{
		\begin{minipage}[c]{0.45\linewidth}
			\label{fig:ReprojectionError}
			\centering
			\includegraphics[width=\linewidth]{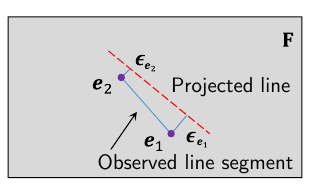}
	\end{minipage}}
	\subfloat[Line \add{triangulation}]{
		\begin{minipage}[c]{.45\linewidth}
			\label{fig:LineTriangulation}
			\centering
			\includegraphics[width=\linewidth]{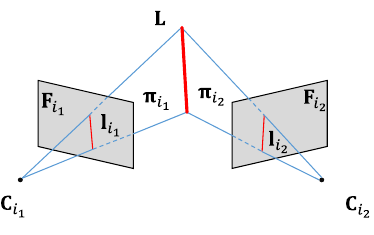}
	\end{minipage}}
	
	\caption{\add{3D line projection and triangulation.}}
	\label{fig:LineSegmentOperation}
\end{figure}

\subsection{Short-term Optimization}\label{Short Term Optimization}
Line flows $\tilde{\mathbf{l}}$  relate a set of 2D line segments to their corresponding 3D line segment. On the other hand, points are important for stable initialization and help in regions with rich texture. Thus, we jointly adopt line flow and 2D-3D point constraints $\mathbf{p}$ to solve camera pose by minimizing an energy function:

\begin{equation}
C_s=\underset{ \mathbf{T}_t }{\min} { \sum_{
\tilde{\{\mathbf{l}\}},\mathbf{\{p\}}}
	}	
\add{\rho(\bm{\epsilon}_\mathbf{l}^T\bm{\mathrm{\Sigma}}^{-1}_{\mathbf{l}} \bm{\epsilon}_\mathbf{l})
	+\rho(\bm{\epsilon}_\mathbf{p}^T\bm{\mathrm{\Sigma}}^{-1}_{\mathbf{p}} \bm{\epsilon}_\mathbf{p})}
,
\label{eq:energy}
\end{equation}
where $\bm{\epsilon}_\mathbf{l}$  and $\bm{\epsilon}_\mathbf{p}$  are the reprojection errors in terms of lines and points, respectively. $\bm{\mathrm{\Sigma}}_\mathbf{p} $ and $\bm{\mathrm{\Sigma}}_\mathbf{l}$ are the covariance matrices for points and line segments, respectively, which are associated with the scales of detected features. \add{We set the covariance matrices of both points and line segments the same as PL-SLAM Mono~\cite{Pumarola2017}.} 
 $\rho(\cdot)$ is a Huber function adopted to increase robustness.

The reprojection error is modelled by following~\cite{Zhang2015}. We define the line reprojection error $\bm{\mathrm{\epsilon}}_\mathrm{l}$ as the distance from the two observed endpoints, $\mathbf{e}_1$ and $\mathbf{e}_2$, to the projected line segment of 3D line $\mathbf{L}$, as illustrated in Fig.~\ref{fig:LineSegmentOperation} The error functions of lines and points are defined as:
\begin{align}
\mathbf{K}_l &= \begin{bmatrix}
f_x & 0 & 0 \\
0 & f_y & 0 \\
-f_yx_0 & -f_xy_0 & f_xf_y
\end{bmatrix}, \mathbf{K}_p = \begin{bmatrix}
f_x & 0 & x_0 \\
0 & f_y & y_0 \\
0 & 0 & 1
\end{bmatrix},\\
\bm{\mathrm{\epsilon}}_\mathbf{l} &=\frac{1}{\sqrt{l_x^2+l_y^2} } \left[\mathbf{e}_1 \vert \mathbf{e}_2\right] ^T  \mathbf{K}_l \begin{bmatrix}
\mathbf{R} \vert \left[\mathbf{t}\right]_\times \mathbf{R} \end{bmatrix} \mathbf{L}, \\
\bm{\mathrm{\epsilon}}_\mathbf{p} &= \mathbf{p} - \lfloor \mathbf{K}_{p} \mathbf{(RP+t) \rfloor}, 
\end{align}
where $\mathbf{K}_l$ is the intrinsic matrix for lines~\cite{Zhang2015}, and $\mathbf{K}_p$ is the intrinsic matrix for points. $f_x$ and $f_y$ are the focal lengths and $(x_0,y_0)^T$ is the principle point. $\left[\mathbf{t}\right]_\times$ is the antisymmetric form of $\mathbf{t}$. $\mathbf{p}$ and $\mathbf{P}$ denote 2D and 3D points, \add{respectively}. Assuming that the projection of 3D line segment $\mathbf{L}$ on a 2D image is $(l_x,l_y,l_z)$, $\frac{1}{\sqrt{l_x^2+l_y^2} }$ is a term used for distance normalization.

\subsection{Long-term Optimization}\label{Long Term Optimization}
The long-term optimization is performed at each keyframe to jointly refine line flows, 3D maps, and camera poses. We follow~\cite{Mur-Artal2017} to manage keyframes. The management of a 3D line map includes 3D line creation, outlier rejection, merging and updating based on line flow representation. We triangulate the latest two line segments within each line flow, reject outliers and merge correlated 3D lines to maintain a reliable line map. We then \add{perform} a local bundle adjustment. Meanwhile, an optional global optimization with loop detection~\cite{Mur-Artal2014} can be performed.

\noindent\textbf{Line flow triangulation.}
 Triangulation is performed when a line flow has survived at least two key frames and the angle between two planes $\pi_i$ and $\pi_j$ is greater than $1^\circ$ (as illustrated in  Fig.~\ref{fig:LineSegmentOperation}~(b)). A 3D line $\mathbf{L}=(\mathbf{n}_L^T,\mathbf{v}_L^T)^T$ is represented by intersecting two planes generated from corresponding 2D line segments and the camera centres:
\begin{equation}
\begin{aligned}
\bm{\pi}_i 	&\leftarrow \left(\mathbf{n}_i,d_i\right) = \begin{bmatrix}\mathbf{R} \mid \mathbf{t} \end{bmatrix}^T \mathbf{K}_l^T \mathbf{l}.\\
\mathbf{L} &\leftarrow \left\{
\begin{aligned}
&	\mathbf{n}_{\mathbf{\mathit{L}}}=d_{i_2} \mathbf{n}_{i_1}- d_{i_1} \mathbf{n}_{i_2}, \\
&	\mathbf{v}_{\mathbf{\mathit{L}}}=\mathbf{n}_{i_2} \times \mathbf{n}_{i_1},
\end{aligned}
\right.
\end{aligned}
\end{equation}

Then, we check the projection errors between the 3D line segment $\mathbf{L}$ and 2D line segment within the corresponding line flow. When the number of outliers is greater than $90\%$ of the number of 2D line segments, we remove the line flow.

We maintain 3D line segment endpoints after successful triangulation. We compute the mean of back-projected 3D endpoints for each 2D line segment as the initial 3D line endpoints. We leverage 2D start/terminal endpoints to calculate 3D start/terminal endpoints. Then, when each new line segment is entered, we update 3D start/terminal endpoints by an averaging operation. The averaging operation is simple and highly efficient because the outlier rejection operation can remove unstable endpoints, and the line flow merging process fuses broken line segments.

\begin{figure}[t]
	\centering
\subfloat[Endpoints uncertainty model]{
	\begin{minipage}[c]{0.45\linewidth}
		\label{fig:Uncertainty}
		\centering
		\includegraphics[width=\linewidth]{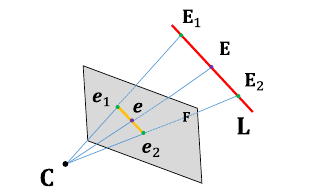}
\end{minipage}}
\subfloat[3D line fusion]{
	\begin{minipage}[c]{.45\linewidth}
		\label{fig:LineFusion}
		\centering
		\includegraphics[width=\linewidth]{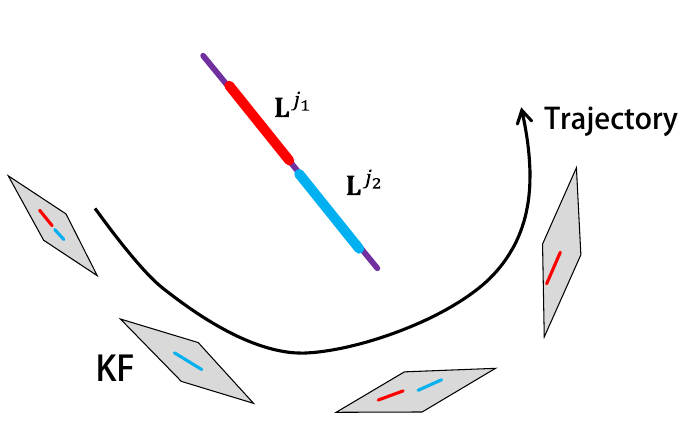}
\end{minipage}}
	
	\caption{\add{Maintaining a 3D line.}}
	\label{fig:LineSegmentOperation_B}
\end{figure}

\noindent\textbf{Outlier rejection.}
As illustrated in Fig.~\ref{fig:LineSegmentOperation_B}~(a), if a 3D line segment is far from the camera, a small disturbance on the image plane will lead to a large deviation, which makes the 3D line segment highly unstable. To handle this issue, we calculate the uncertainty  $\mathrm{U}_{\mathbf{e}}$ according to the distance between 2D and 3D line segment endpoints:
\begin{equation}
\mathrm{U}_{\mathbf{e}} =\frac{ \left\|\mathbf{E}_1 - \mathbf{E}_2\right\| } { \left\|\mathbf{e}_1 - \mathbf{e}_2\right\| },
\end{equation}
where the 2D pixels $\mathbf{e}_1$ and $\mathbf{e}_2$ are 0.5 pixels away from $\mathbf{e}$ along the line direction. $\mathbf{E}_1$ and $\mathbf{E}_2$ are the  resulting projected 3D points.

Larger uncertainty implies more unstable endpoints. In practice, we update the mean $\mu_\mathrm{U}$ and standard deviation $\sigma_\mathrm{U}$ of all line segment endpoints every 10 key-frames, and reject a 2D line segment if the uncertainty of at least one endpoint exceeds $\mu_\mathrm{U}+3\sigma_\mathrm{U}$. We further find that erroneous line matching tends to generate long 3D line segments. Meanwhile, a 3D line on a plane parallel to the optical axis has unstable endpoints, and is usually observed as a short line segment. Hence, our uncertainty criteria can reject these two kinds of 3D lines to guarantee high reliability.

\noindent\textbf{Line flow merging.} As illustrated in Fig.~\ref{fig:LineSegmentOperation_B}~(b), a 3D line might be visible again after the end of the corresponding line flow due to certain reasons, such as occlusion, out of sight, and severe blurring. These two 3D lines tied to two separate line flows should be merged. We merge two 3D lines $\mathbf{L}^{j_1}$ and $\mathbf{L}^{j_2}$ with their line flows if they are collinear and overlapping.
Assuming that the number of 2D observed line segments of $\mathbf{L}^{j_1}$ is larger than that of $\mathbf{L}^{j_2}$, we merge the set of 2D observations of $\mathbf{L}^{j_2}$ into the set of observations of $\mathbf{L}^{j_1}$.
3D endpoints are updated by computing the mean of back-projected 3D endpoints within the new set of 2D line segments.

\noindent\textbf{Local bundle adjustment.}
Local bundle adjustment is performed by minimizing the reprojection errors of points and line segments:
\begin{equation}
\mathrm{E}= \underset{\mathbf{T}_\varphi,\{\mathbf{L}\},\{\mathbf{P}\}}{{\min }}
\sum_{\tilde{\{\mathbf{l}\}},\{\mathbf{p}\}} \add{\rho(\bm{\epsilon}_\mathbf{l}^\mathrm{T}\bm{\mathrm{\Sigma}}_\mathbf{l}^{-1} \bm{\epsilon}_\mathbf{l})
+\rho(\bm{\epsilon}_\mathbf{p}^\mathrm{T}\bm{\mathrm{\Sigma}}_\mathbf{p}^{-1} \bm{\epsilon}_\mathbf{p})},
\label{eq:OPT}
\end{equation}
where we optimize the camera poses $\mathbf{T}_\varphi$, line segments $\mathbf{L}$ and points $\mathbf{P}$ in the covisibility graph. \add{All the keyframes in the covisibility graph are close in the 3D space.
For simplicity, when we process the current keyframe, the camera poses $\mathbf{T}_\varphi$ include only those of the current keyframe and its connected keyframes in the local graph. Two keyframes are connected when they share sufficient 3D features. The other keyframes are included in the optimization but remain fixed.}

We check the line reprojection errors within the line flow. When the number of outliers is larger than half of the total number of line segments within the line flow, we delete the line flow. In addition, we re-compute the mean of 3D back-projected endpoints from 2D line segments to update 3D endpoints.

\noindent\textbf{Loop closure.}
When we detect a loop closure using the method in~\cite{Mur-Artal2014},
we update all the keyframes and the map with a similarity transformation to eliminate scale drifts. A two-step strategy is implemented to accelerate the convergence. First, following~\cite{Mur-Artal2014}, we obtain observed pose-pose constraints with $\mathbf{T}^o_{i,j}$, $\mathbf{T}_{i,j} \in \mathbf{T}_{\phi} $ to perform a pose graph optimization:
\begin{align}
\mathrm{E}&=\underset{\mathbf{T}_\phi}{\min}
\sum_{\mathbf{T}_{i,j} \in \mathbf{T}_{\phi}} \add{\bm{\epsilon}_{\mathbf{T}_{i,j}}^\mathrm{T}
\bm{\mathrm{\Sigma}}_{\mathbf{T}_{i,j}}^{-1}
\bm{\epsilon}_{\mathbf{T}_{i,j}} }, \\
\bm{\epsilon}_{\mathbf{T}_{i,j}} & = \mathbf{T}^o_{i,j} \mathbf{T}_j \mathbf{T}_i^{-1}.
\end{align}
To address the problem of scale drifts, we transform the 3D line $\mathbf{L}$ from the world coordinate to the reference keyframe coordinates before optimization. Then, the parameters of all the keyframes and the 3D map are jointly optimized in the similarity transformation:
\begin{align}
\mathrm{E}&= \underset{\{\mathbf{T}\},\{\mathbf{L}\},\{\mathbf{P}\}}{{\min }}
\sum_{\{\tilde{\mathbf{l}}\},\{\mathbf{p}\}} \add{\rho(_{s}\bm{\epsilon}_\mathbf{l}^\mathrm{T}\bm{\mathrm{\Sigma}}_\mathbf{l}^{-1} {_{s}\bm{\epsilon}}_\mathbf{l})
+\rho({_{s}\bm{\epsilon}}_\mathbf{p}^\mathrm{T}\bm{\mathrm{\Sigma}}_\mathbf{p}^{-1} {_{s}\bm{\epsilon}}_\mathbf{p})}
,\\
_{s}\bm{\mathrm{\epsilon}}_\mathbf{l} &=\frac{1}{\sqrt{l_x^2+l_y^2} } \left[\mathbf{e}_1 \vert \mathbf{e}_2\right] ^T  \mathbf{K}_l \begin{bmatrix}
\mathbf{R} \vert s\left[\mathbf{t}\right]_\times \mathbf{R} \end{bmatrix} \mathbf{L}, \\
_{s}\bm{\mathrm{\epsilon}}_\mathbf{p} &= \mathbf{p} - \lfloor s\mathbf{K}_{p} \mathbf{(RP+t) \rfloor}.
\end{align}
After this optimization, the same procedure is performed in the local bundle adjustment module to discard outliers.

\section{Experiments} \label{Sec:Experiment}

We compare our approach with state-of-the-art SLAM systems on indoor and outdoor datasets, including the TUM RGBD~\cite{Sturm2012}, the 7-Scenes~\cite{Shotton2013}, the EuROC~\cite{Burri2016} and the KITTI~\cite{Geiger2012} datasets. We only use the monocular RGB images in these datasets. Both accuracy and efficiency are evaluated to show the superiority of the proposed LF-SLAM. We also provide 3D line maps and line flow tracking results as qualitative evaluations to demonstrate that reliable and stable line flows can be extracted in challenging scenarios.

\subsection{Implementation Details}
All the experiments are performed on a desktop PC with a 3.6 GHz Core i7-7700 CPU and 16 GB  memory. We use the evo tool~\cite{grupp2017evo} to align metrics and eliminate scale ambiguity before conducting quantitative comparisons.

We utilize the Levenberg Marquardt algorithm implemented in the Ceres solver~\cite{ceres-solver} for solving nonlinear least squares problems. In the appendix, we discuss line representation and the Jacobian matrices of error terms used in the optimization procedure.

\noindent\textbf{Parameter setting.}
We discard line segments when the length is shorter than 0.005 of the image diagonal~\cite{Hofer2017}. \add{We merge two line segments in 2D or 3D spaces once they satisfy the collinearity condition, i.e., the angular between the two lines is less than a threshold.  A small threshold, e.g., 2D: $ <3^\circ$ and 3D: $<5^\circ$, leads to collinearity check failure. A large threshold, e.g., 2D: $ > 10^\circ$ and 3D: $ > 15^\circ$, causes wrongly merging.
Considering the trade-off between recall and precision, we set the 2D line collinearity threshold to $5^\circ$; and that for the 3D line collinearity to $10^\circ$.}

\add{A line flow allows $\alpha$ successive frames to have no observations, considering temporary occlusion. A large $\alpha$ risks wrongly grouping of different line flows. Here, we set $\alpha$ to 3.}

The width of the rectangular search area illustrated in
Fig.~\ref{fig:LineFlowPrediction} (b) is determined according to the previous support region and 2D
motions. \add{Then $5\times5$ seeds are uniformly sampled within the
rectangle. A sparse grid with less seeds leads to fewer line candidates, while a dense one results in too many candidates and burdens the following computations.}

\add{As to the size of the temporal window for line prediction, i.e., $t_w$, it is feasible to estimate camera motion with $t_w=2$. However, the velocity estimation using only two frames is not stable. Therefore, we set $t_w=5$ for a more stable velocity. In our experiments, window sizes ranging from $3\sim7$ bring similar performance.}

\subsection{Line Flow Analysis}
\begin{figure}[t]
	\centering
	\subfloat[Updating time (ms)]{
		\begin{minipage}[c]{.45\linewidth}
			\centering
			\includegraphics[width=\linewidth]{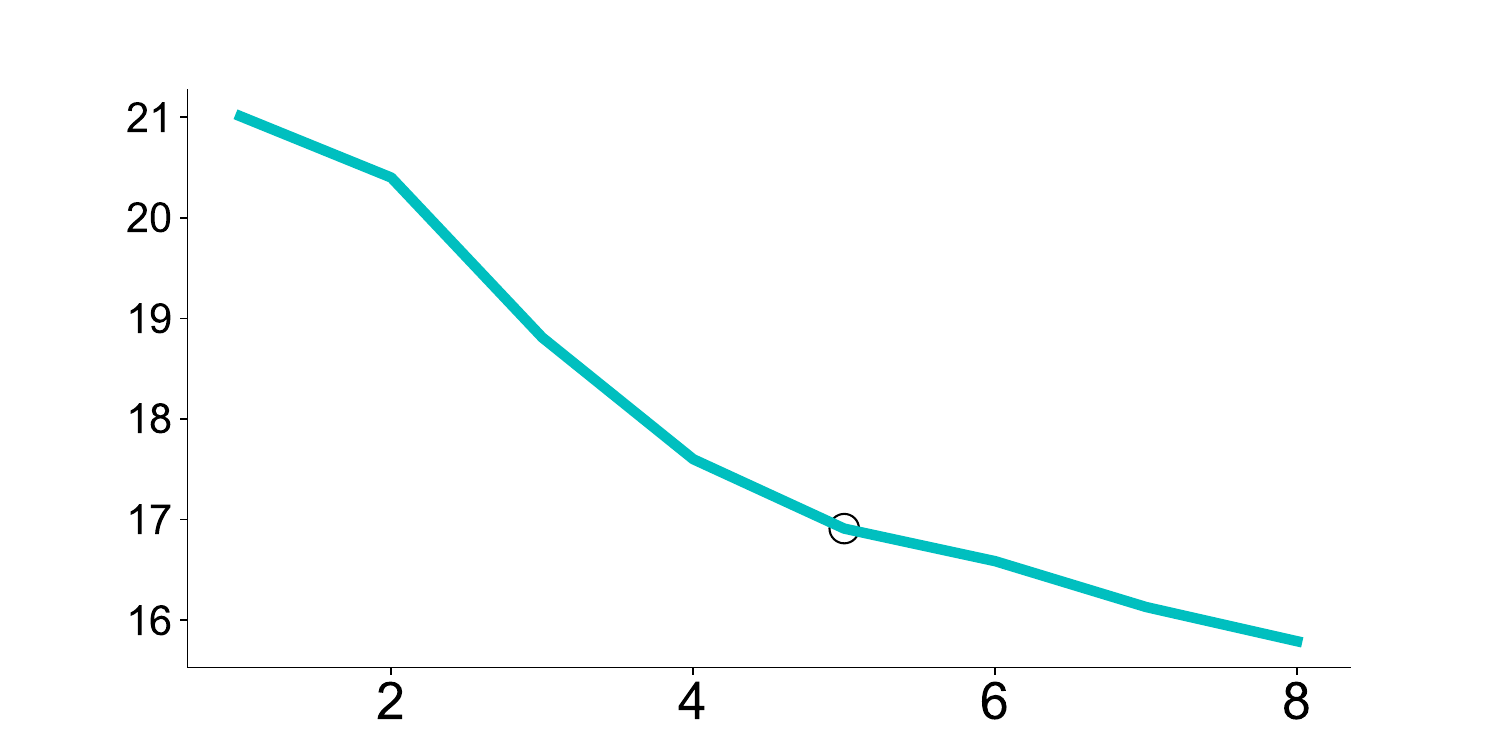}
		\end{minipage}
		\label{subfig:nclTime}
	}
	\hfil
	\subfloat[3D Line numbers]{
	\begin{minipage}[c]{.45\linewidth}
		\centering
		\includegraphics[width=\linewidth]{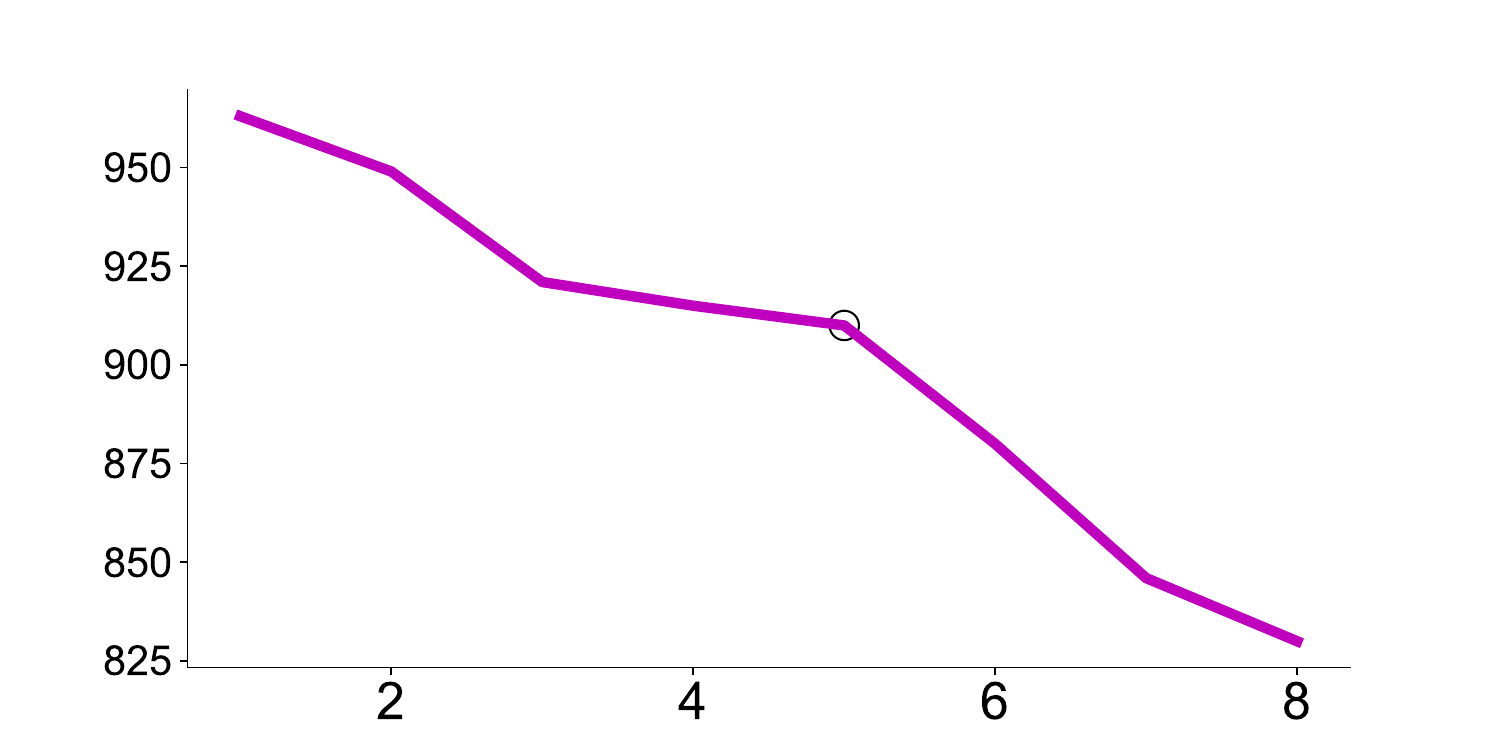}
	\end{minipage}
		\label{subfig:ncl3DLine}
	}
	\vfil
	\subfloat[Mean translation error (cm/s)]{
		\begin{minipage}[c]{.45\linewidth}
			\centering
			\includegraphics[width=\linewidth]{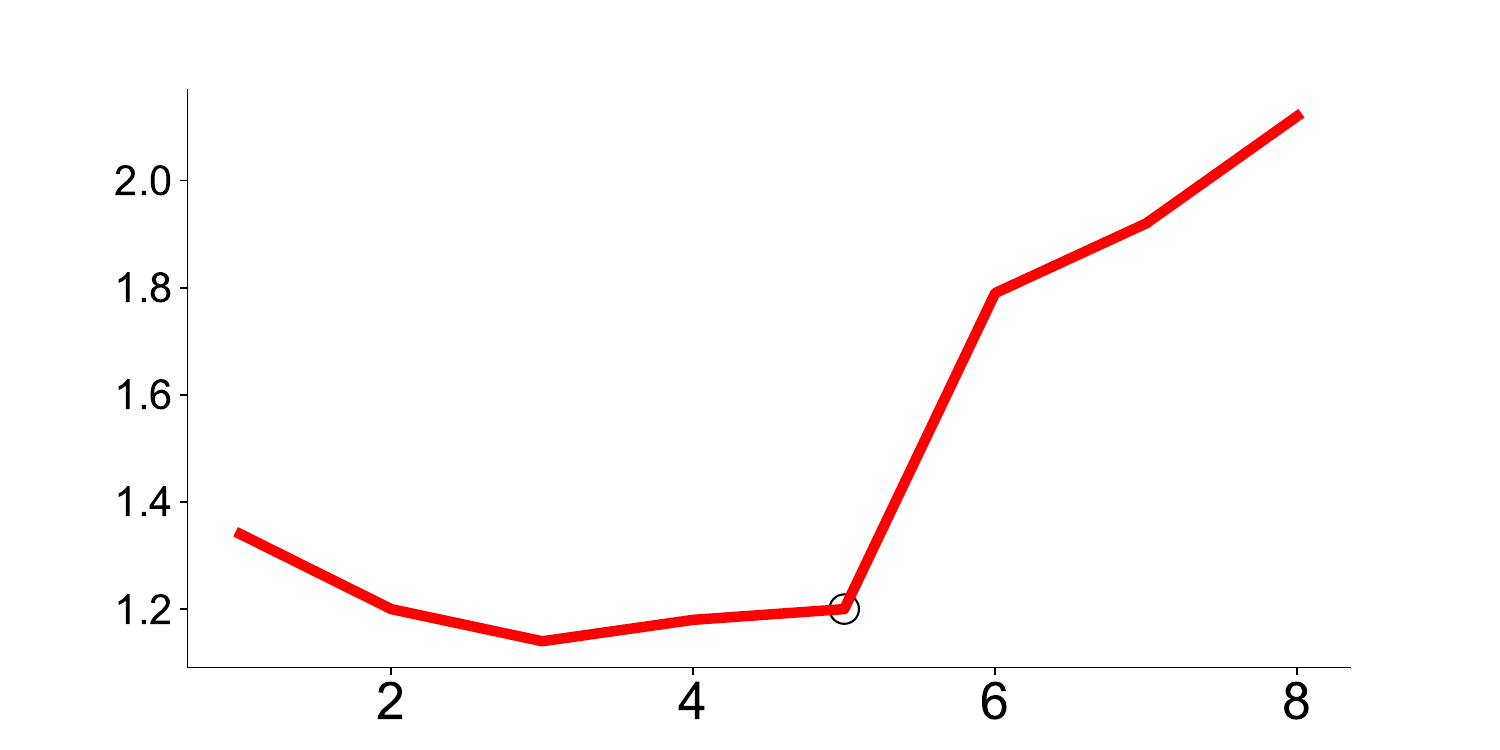}
		\end{minipage}
		\label{subfig:nclMTE}
	}
	\hfil
	\subfloat[Mean rotation error (deg/s)]{
		\begin{minipage}[c]{.45\linewidth}
			\centering
			\includegraphics[width=\linewidth]{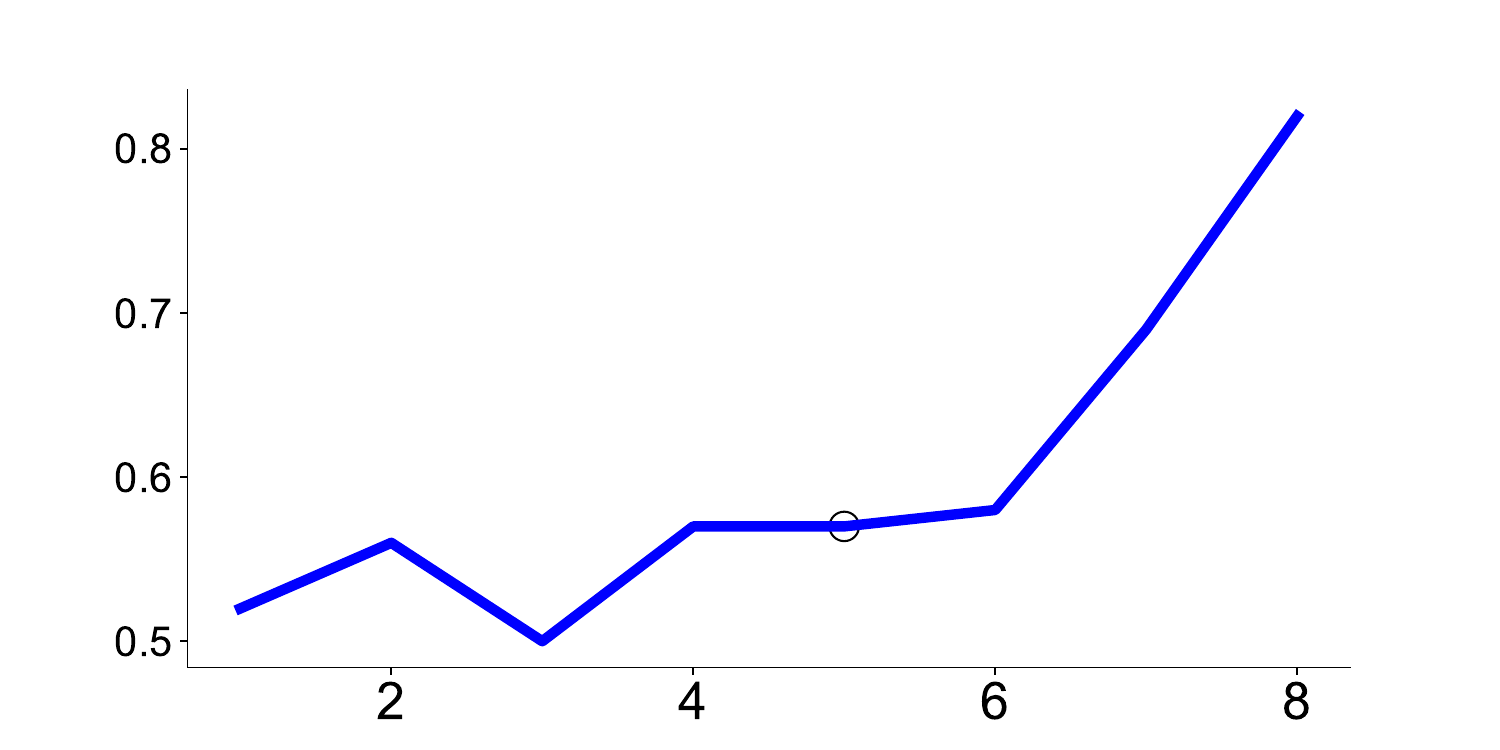}
		\end{minipage}
		\label{subfig:nclMRE}
	}
	
	\caption{The influence of $N_{cl}$ on updating time, 3D line number, mean translation error, and rotation error. The experiment is conducted on \textit{fr3\_long\_office} sequence.}
	\label{fig:LineFlowAnalysis}
\end{figure} 

\noindent\textbf{Key parameter study.} $N_{cl}$ determines the frequency of the running of an LSD for the remaining pixels in a frame after the pixels in the predicted regions are visited. $N_{cl}$ is a very important parameter since it  determines the efficiency and effectiveness of line flow extraction. We conducted a comprehensive study to understand the effect of $N_{cl}$ on efficiency and robustness (shown in Fig.~\ref{fig:LineFlowAnalysis}). A large $N_{cl}$ leads to high efficiency. However, a very large $N_{cl}$ increases pose errors. To balance the trade-off between efficiency and robustness, we set $N_{cl}$ to 5 in all the experiments.

\begin{figure}[t]
	\centering
	\subfloat[RBG image]{
			\includegraphics[width=.45\linewidth]{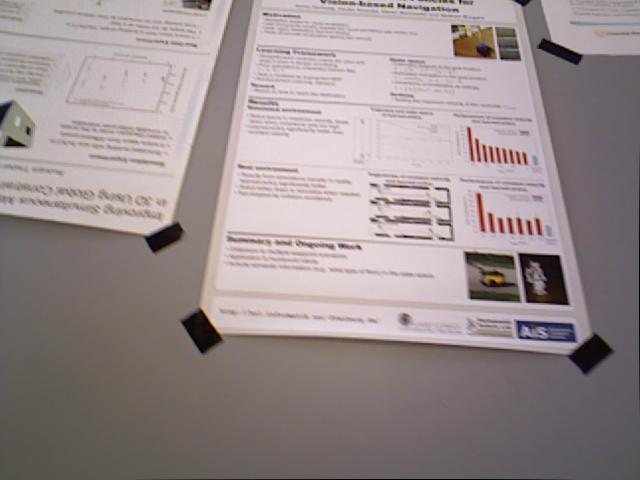}
	}
	\subfloat[Visited pixels]{
		\includegraphics[width=.45\linewidth]{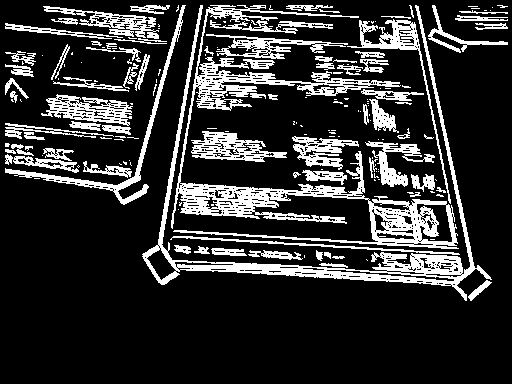}
	}

	\caption{ Visualization of visited pixels by guided line detection in the \textit{fr3\_nostr\_tex\_near} sequence. (a) An $640 \times 480$ image; (b) shows The corresponding pixels (white) needed to be processed.}
	\label{fig:LineFlowTheory}
\end{figure}

\begin{figure*}[t]
	\centering
	\begin{minipage}[c]{\linewidth}
		\centering
		\includegraphics[width=.95\linewidth]{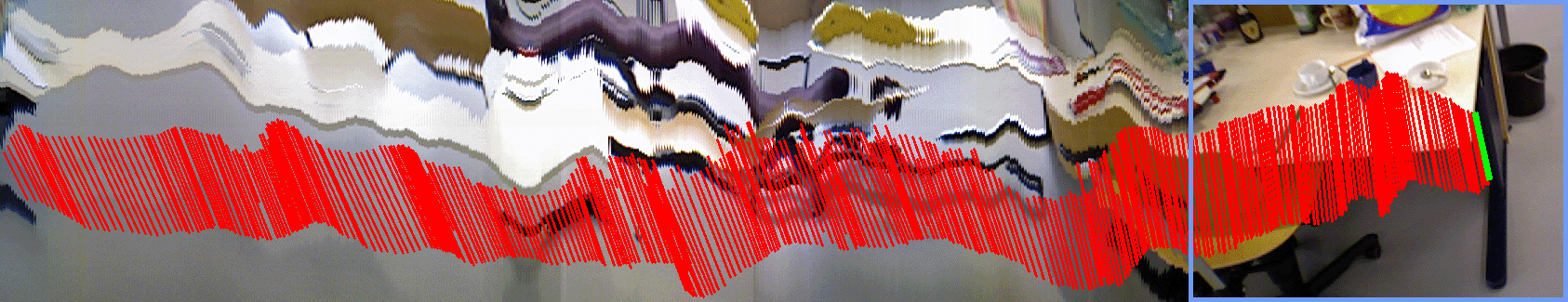}
	\end{minipage}
	\caption{Line flow visualization. We display the tracking result of a line flow over 300 frames on the \textit{fr3\_long\_office} sequence. Green line segments are in current frame (blue) while red line segments are in previous ones. 
	}
	\label{fig:LineFLowVisualization}
\end{figure*}

\noindent\textbf{Ablation study.}
We use LF-P-SLAM and LF-PL-SLAM to denote two variants of the proposed LF-SLAM. LF-P-SLAM is our baseline with only points. LF-PL-SLAM adds lines such as PL-SLAM Mono, namely using LSD and LBD algorithms for line extraction and description, respectively, but treats both 3D line management and optimization as LF-SLAM. Table~\ref{table:EuRoCMAV} shows the absolute pose errors (ATEs) of keyframe trajectories on the EuRoC dataset. From the table, it can be seen that the results of LF-PL-SLAM are better than those of LF-P-SLAM on 6 sequences. LF-SLAM has lower ATEs than LF-P-SLAM on all sequences. LF-SLAM performed better than LF-P-SLAM on 8 sequences. For robustness, LF-P-SLAM fails on 2 difficult sequences, LF-PL-SLAM fails on 1 difficult sequence, and LF-SLAM  successfully runs all sequences.

\noindent\textbf{Running time.}
To compare the efficiency of different line segment extraction and matching strategies, we first test the running time using LSD~\cite{VonGioi2010} for extraction and LBD~\cite{Zhang2013} for matching. This strategy has been adopted by a few SLAM works~\cite{Gomez-Ojeda2017,Pumarola2017,Zhao2018_ECCV}.
 It takes 37.97 ms for each frame.  As illustrated in Fig.~\ref{fig:LineFlowAnalysis}~(a), setting $N_{cl}$ to 1, the module can be viewed as LSD + line flow prediction implementation and runs in 21.01 ms. In comparison, our line flow predicting and updating implementation takes 16.91 ms because we only process selected pixels. The guided line extraction and descriptor-free matching ensure high efficiency. We illustrate the visited pixels of an image when we utilize the line flow updating strategy in Fig.~\ref{fig:LineFlowTheory}.
 This strategy guarantees that line flow based SLAM runs in real time ($25-35$ FPS).
The time costs of the different modules of LF-SLAM are given in Table~\ref{table:TimeAnalysis}.
Line flow tracking  is  a joint procedure for feature extraction and 2D-2D matching.
The module of line flow tracking runs at 70 FPS on the TUM-RGBD dataset and at 50 FPS on the EuRoC MAV, because of the high  image resolution of the EuRoC MAV dataset.
The point tracking and line flow tracking run in parallel.
The front-end tracker (including line flow tracking, point tracking, and short-term optimization) takes 28 ms/frame on the TUM-RGBD and EuRoC MAV datasets.
We put the long-term optimization in another thread to achieve real time performance. Note that  3D-2D line matching is automatically performed in line flows based on the line flow definition. Line merging is very efficient because most of the correspondences for line segments are already found and kept in line flows.

\setlength{\tabcolsep}{5pt}
\begin{table}[!t]
	\centering
	\small
	\captionsetup{labelsep=newline}
	\captionsetup{justification=centering}
	
	\caption{{\MakeUppercase{The Average Runtime of Each module of LF-SLAM} (ms).}}
	\begin{tabular}{*{5}{l}}
		\hline
		\hline
		&  & & TUM-RGBD & EuRoC MAV\\
		Resolution &  & & 640 $\times$ 480 & 752 $\times$ 480 \\
		\hline
		\multicolumn{3}{l}{Line Flow Tracking} &\multicolumn{1}{r}{14.29 } & \multicolumn{1}{r}{19.93 } \\
		\multicolumn{3}{l}{Point Tracking} & \multicolumn{1}{r}{23.98 } & \multicolumn{1}{r}{25.59 } \\
		\multicolumn{3}{l}{Short-term Opt.} & \multicolumn{1}{r}{3.69 } & \multicolumn{1}{r}{3.01 }\\
		\hline
		\multicolumn{1}{l}{\multirow{4}*{Long-term Opt.}} & \multicolumn{2}{l}{3D Line Proc.} & \multicolumn{1}{r}{0.83 } & \multicolumn{1}{r}{1.74 } \\
		&  \multicolumn{2}{l}{3D Point Proc.} & \multicolumn{1}{r}{25.55 } & \multicolumn{1}{r}{23.30 } \\
		&  \multicolumn{2}{l}{Local BA} & \multicolumn{1}{r}{257.80 } & \multicolumn{1}{r}{301.28 } \\
		&  \multicolumn{2}{l}{Loop Closure} & \multicolumn{1}{r}{1164.66 } & \multicolumn{1}{r}{289.25 } \\
		\hline
		\hline
	\end{tabular}
	\label{table:TimeAnalysis}
\end{table}

\begin{figure}[!t]
	\centering
	\subfloat[7-Scenes. \textit{fire\_02}]{
		\label{subfig:occ}
		\begin{minipage}[c]{0.45\linewidth}
			\centering
			\includegraphics[width=\linewidth]{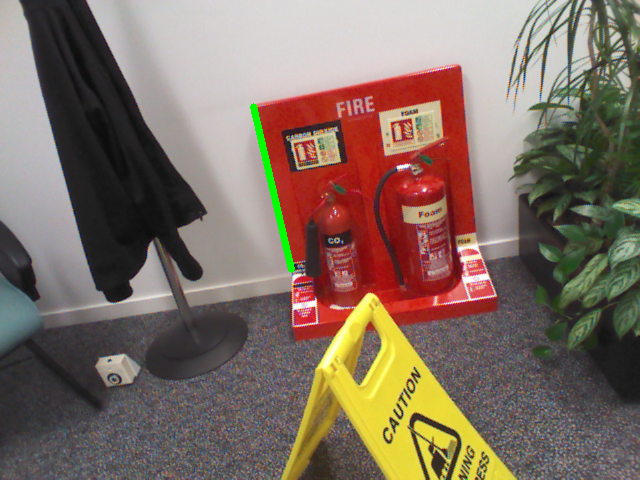}
		\end{minipage}
		\space
		\begin{minipage}[c]{.45\linewidth}
			\centering
			\includegraphics[width=\linewidth]{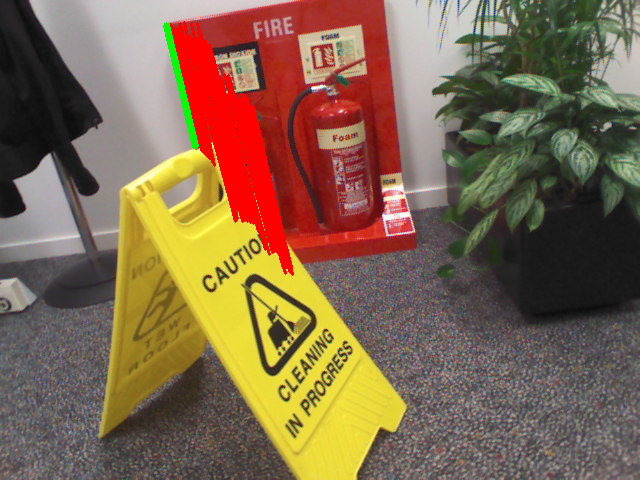}
		\end{minipage}
	}
	\vfil
	\subfloat[7-Scenes. \textit{stairs\_04}]{
		\label{subfig:rep}
	\begin{minipage}[c]{0.45\linewidth}
		\centering
		\includegraphics[width=\linewidth]{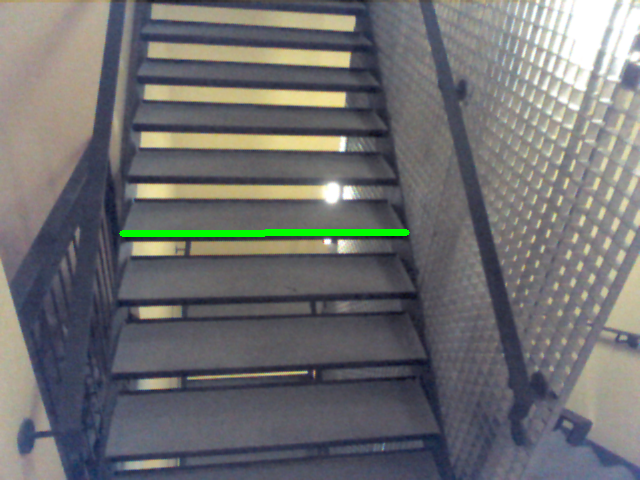}
	\end{minipage}
	\space
	\begin{minipage}[c]{.45\linewidth}
		\centering
		\includegraphics[width=\linewidth]{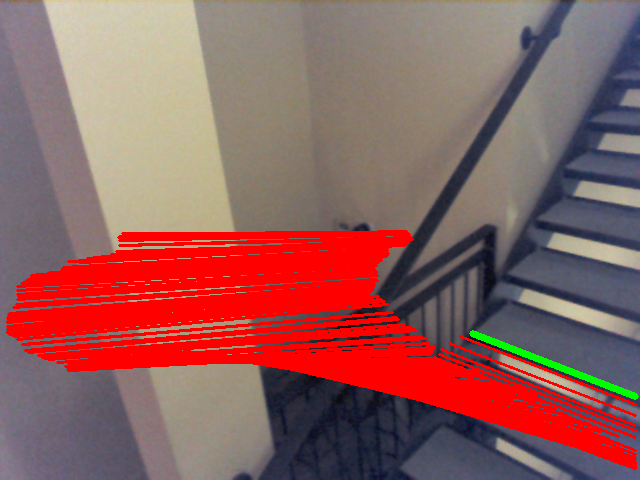}
	\end{minipage}
	}
	\vfil
	\subfloat[TUM. \textit{fr2\_desk\_with\_person}]{
		\label{subfig:dyn}
	\begin{minipage}[c]{0.3\linewidth}
		\centering
		\includegraphics[width=\linewidth]{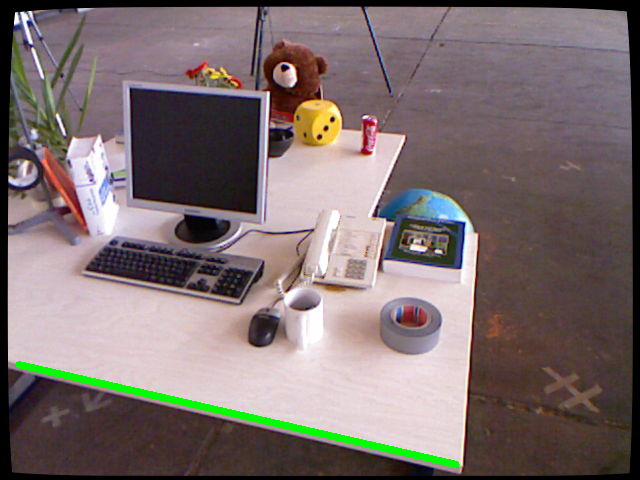}
	\end{minipage}
	\space
	\begin{minipage}[c]{.6\linewidth}
		\centering
		\includegraphics[width=\linewidth]{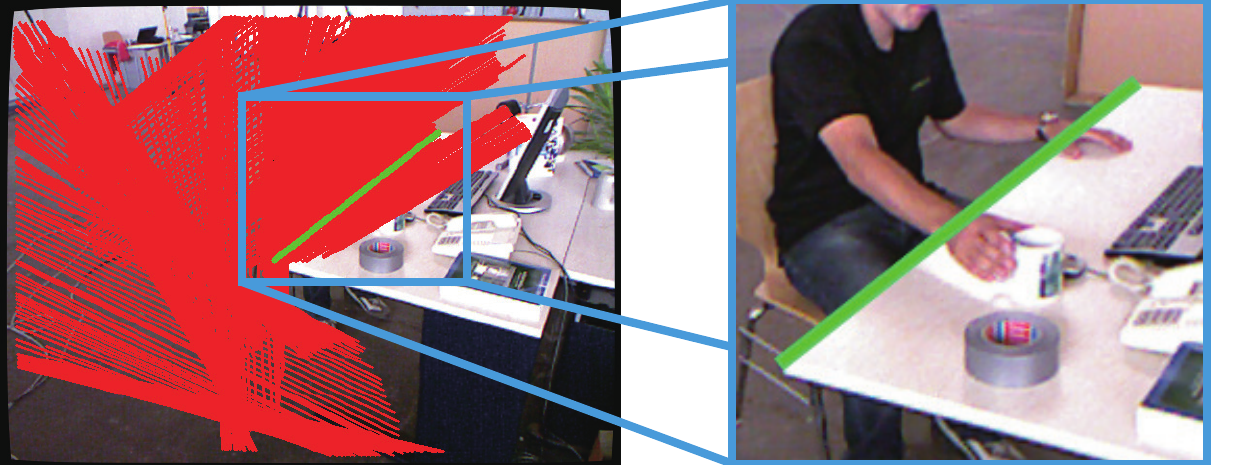}
	\end{minipage}
	}
	\caption{Line flow results in 3 challenging scenarios: partial occlusion (a), repetitive texture (b), and dynamic object occlusion (c). The left picture of each row shows the first line segment of the line flow (green), while the right one shows the current line segment of the line flow (green) along with previously extracted line segments (red).}
	\label{Fig:Reslut}
\end{figure}

\noindent\textbf{Line flow visualization.} We visualize a line flow on the \textit{fr3\_long\_office} sequence in Fig.~\ref{fig:LineFLowVisualization}. Starting from the first frame, we show all the 2D line segments. The line segments move with the camera motion. At the same time, the endpoints and line direction are stable on each frame from different views.

\setlength{\tabcolsep}{12pt}
\begin{table*}[t]
	\centering
	\small
	\captionsetup{labelsep=newline}
	\captionsetup{justification=centering}
	\caption{\MakeUppercase{The absolute keyframe trajectory error (ATE) evaluation on the TUM RGBD  benchmark}~\cite{Sturm2012} (RMSE, cm).
		\MakeUppercase{The results of PL-SLAM Mono (Origin)  and LSD-SLAM are derived from}~\cite{Pumarola2017}. \MakeUppercase{The result of ORB-SLAM2 are derived from}~\cite{Mur-Artal2017}. \MakeUppercase{The result of DSO$^\dagger$}~\cite{Engel2018} \MakeUppercase{is generated from the source code with the default parameters.}
	}
	\begin{threeparttable}
	\begin{tabular}{*{8}{c}}
		\hline
		\hline
		\multicolumn{2}{c}{\multirow{2}*{\diagbox{Dataset}{Method}}} &
		\multicolumn{1}{c}{\multirow{2}*{LF-SLAM}}&
		\multicolumn{2}{c}{\multirow{1}*{{PL-SLAM Mono~\cite{Pumarola2017}}}} &
		\multicolumn{1}{c}{\multirow{2}*{{ORB-SLAM2~\cite{Mur-Artal2017}}}} & 
		\multicolumn{1}{c}{\multirow{2}*{{DSO$^\dagger$~\cite{Engel2018} }}} & 
		\multicolumn{1}{c}{\multirow{2}*{{ LSD-SLAM~\cite{Engel2014} }}} \\
		&& & Re-imp & Ori & & &\\
		\hline
		\multicolumn{2}{l}{fr1\_xyz} & 1.05 &1.21& 1.21 & \textbf{0.90} & 6.30 & 9.00 \\
		\multicolumn{2}{l}{fr1\_floor}  &\textbf{1.74}&3.91 &7.59& 2.99  & 5.25 & 38.07 \\
		\multicolumn{2}{l}{ fr2\_xyz} &\textbf{0.25} &0.42 &0.43 &0.30  & 0.98& 2.15 \\
		\multicolumn{2}{l}{ fr2\_360\_kidnap} &\textbf{2.97} &4.60 &3.92&3.81 &4.12 & - \\
		\multicolumn{2}{l}{fr2\_desk\_with\_person} &\textbf{0.69} &1.49 &1.99  & 0.88 & - & 31.73\\
		\multicolumn{2}{l}{fr3\_str\_tex\_far}& 0.88&1.00& 0.89 &\textbf{0.77}& 1.36 & 7.95 \\
		\multicolumn{2}{l}{fr3\_str\_tex\_near}& \textbf{1.17}&1.48&1.25& 1.58& 7.26 & - \\
		\multicolumn{2}{l}{fr3\_nostr\_tex\_near}& \textbf{1.36}&1.39 & 2.06& 1.39& 7.30  & 7.54 \\
		\multicolumn{2}{l}{fr3\_sit\_halfsph} & \textbf{1.29} &1.91& 1.31 & 1.34 & 3.57 & 5.87 \\
		\multicolumn{2}{l}{fr3\_long\_office}&\textbf{1.35}&1.40  &1.97& 3.45 & 10.11 & 38.53  \\
		\multicolumn{2}{l}{fr3\_walk\_xyz} & \textbf{1.16}&1.38  & 1.54 & 1.24 &  14.14 & 12.44  \\
		\multicolumn{2}{l}{fr3\_walk\_halfsph} & 1.66 & 1.40&  \textbf{1.60} & 1.74 &  31.86 & -  \\	
		\hline
		\multicolumn{2}{l}{average} & \textbf{1.29} &1.80& 2.15 & 1.70 & 8.39 & 17.03\\
		\hline
		\hline
		\end{tabular}
	\footnotesize{}
	\end{threeparttable}
	\label{table:ATE}
\end{table*}

In addition, we \add{visualize our line flows} in 3 typical scenarios. Fig.~\ref{Fig:Reslut}~(a) demonstrates that we can track a line segment with stable endpoints \add{across frames with} partial occlusions.
Fig.~\ref{Fig:Reslut}~(b) shows that we can obtain temporally consistent line segments in scenarios with repetitive textures. \add{Fig.~\ref{Fig:Reslut}~(c) demonstrates that the table edge can be detected completely even when it is partially occluded by the human hands.}

\subsection{Quantitative Comparison in Localization}

\noindent\textbf{Quantitative Evaluation Baselines.} We evaluate LF-SLAM against a few state-of-the-art SLAM algorithms: ORB-SLAM2~\cite{Mur-Artal2017}, PL-SLAM~\cite{Pumarola2017}, LSD-SLAM~\cite{Engel2014} and DSO~\cite{Engel2018}. DSO~\cite{Engel2018} and LSD~\cite{Engel2014} are direct methods that not only operate well in texture-less environments but also provide visually appealing reconstruction results.
Note that PL-SLAMs have monocular~\cite{Pumarola2017} and stereo~\cite{Gomez-Ojeda2017} versions in the literature, and we label the monocular version as PL-SLAM Mono~\cite{Pumarola2017}.
Both ORB-SLAM2~\cite{Mur-Artal2017} and PL-SLAM  Mono~\cite{Pumarola2017} are indirect methods. ORB-SLAM2 has achieved state-of-the-art performance on many datasets. PL-SLAM Mono exploits point and line features to address low texture scenarios.
We reimplement PL-SLAM Mono for more sufficient comparisons, especially in 3D mapping  as there are no official published source codes.
Due to the multithreading interleaving, for each sequence, we run all the systems 10 times and report median values for the trajectory results.

\noindent\textbf{Experiments on the TUM RGBD benchmark.} A quantitative evaluation of localization accuracy on the TUM RGBD benchmark is demonstrated in Table~\ref{table:ATE}. The
TUM RGBD dataset consists of 39 indoor sequences captured in an office environment and an industrial hall.
Indirect methods, e.g., ORB-SLAM2, PL-SLAM Mono, and LF-SLAM, provide better performance than direct methods, such as DSO and LSD-SLAM.
To show that the reimplemented version of PL-SLAM achieves comparable performances with the original, we provide the results of the two versions in Table~\ref{table:ATE}.
 \add{Direct methods perform poorly in localization on the TUM RGBD dataset, which is recorded with a rolling shutter camera and not friendly to direct approaches.}
 LF-SLAM achieves the best results on 9 of 12 sequences. The mean ATEs of our method compared against other methods prove that the LF-SLAM system exhibits good performance in different scenarios.

The \emph{fr1\_floor} sequence contains a few knotholes, which can be easily tracked by point features. In contrast, salient line features rarely appear in this sequence. For this reason, the results of PL-SLAM	Mono are inferior to those of ORB-SLAM2. By carefully maintaining line flows, LF-SLAM achieves the best results by utilizing temporarily visible lines on objects and repetitive textures on the floor.

\begin{figure}[t]
	\centering
	\subfloat[RGB image]{
		\begin{minipage}[c]{.4\linewidth}
			\centering
			\includegraphics[width=\linewidth]{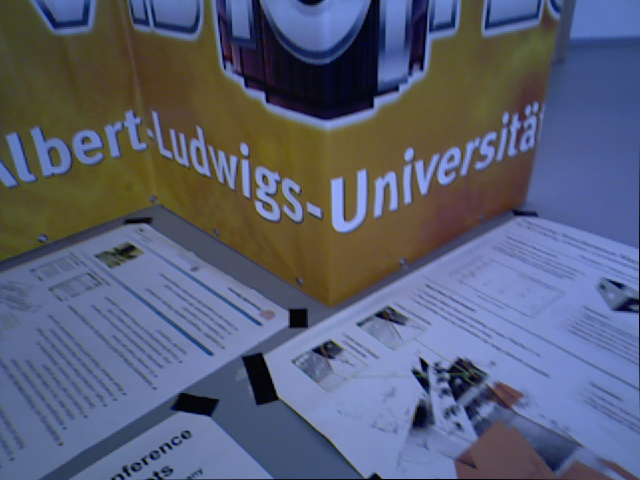}
		\end{minipage}
		\label{fr3_RGB}
	}
	\hfil
	\subfloat[ORB-SLAM]{
		\begin{minipage}[c]{.4\linewidth}
			\centering
			\includegraphics[width=\linewidth]{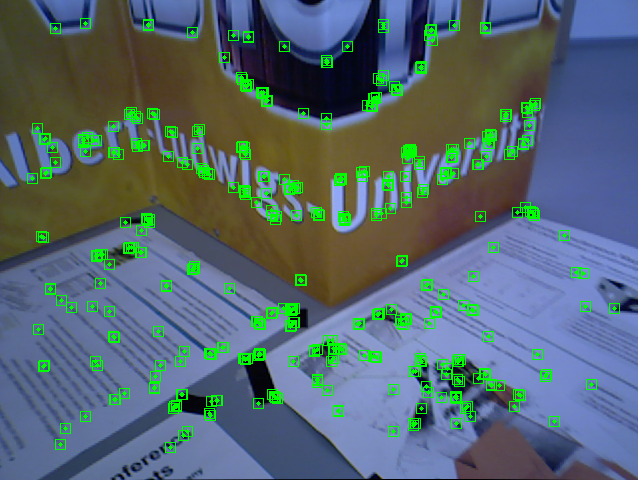}
		\end{minipage}
		\label{fr3_ORB}
	}
	\hfil
	
	\subfloat[PL-SLAM Mono]{
		\begin{minipage}[c]{.4\linewidth}
			\centering
			\includegraphics[width=\linewidth]{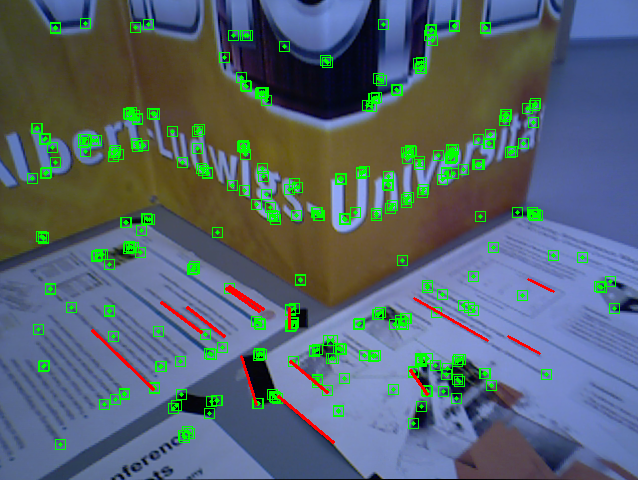}
		\end{minipage}
		\label{fr3_PL-SLAM}
	}
	\hfil
	\subfloat[LF-SLAM]{
		\begin{minipage}[c]{.4\linewidth}
			\label{subfig:fusion}
			\centering
			\includegraphics[width=\linewidth]{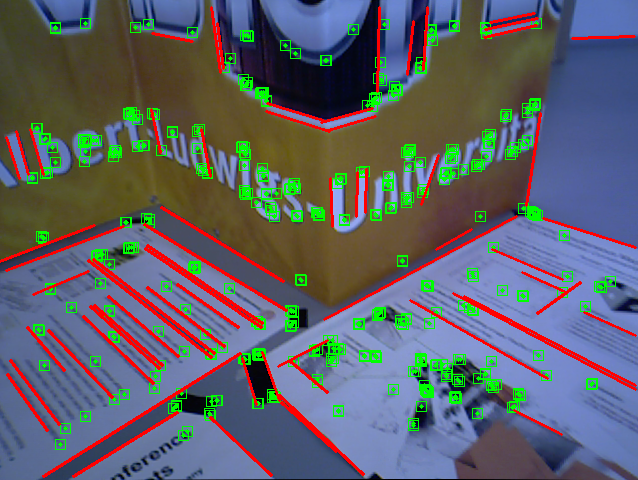}
		\end{minipage}
		\label{fr3_LF-SLAM}
	}
	\caption{\add{We visualize the 3D features reconstructed from the blurred sequence \textit{fr3\_str\_tex\_near} by projecting them on one frame~(a). (b) (c) and (d) show results, points in green and lines in red, obtained by ORB-SLAM, PL-SLAM and the proposed LF-SLAM, respectively.}
	}
	\label{fig:Fr3_result}
\end{figure}

The \emph{fr3\_str\_tex\_near} sequence has abundant texture. However, frequent camera jitters at close range lead to motion blurs. \add{Line flows are stable across blurred frames. As seen from Fig.~\ref{fig:Fr3_result}, many corner points are blurred and failed to be detected. Most of the lines are visually identifiable. Compared to PL-SLAM, our LF-SLAM captures richer and more reliable line segments in these scenes. This gurratees that LF-SLAM performs beyond all the others in localization, as it can be seen in Table~\ref{table:ATE}}.

The \emph{fr3\_long\_office} contains desks and has a large loop closure. Both PL-SLAM Mono and LF-SLAM achieve much better results than ORB-SLAM2. This proves the advantage brought by the integration of line and point features.


\begin{table}[!t]
	\setlength{\tabcolsep}{6pt}
	\centering
	\small
	\captionsetup{labelsep=newline}
	\captionsetup{justification=centering}
	
	\caption{\MakeUppercase{The ATE evaluation on the 7-Scenes}~\cite{Shotton2013} (RMSE, cm). \MakeUppercase{For each scene, Four sequences are used for evaluation. ``-'' denotes that the system fails in the sequence. The results of ORB-SLAM$^\dagger$ are generated from the officially released source code with tunned parameters.}}
	\begin{threeparttable}
	\begin{tabular}{lrrr}
		\hline
		\hline
		\multicolumn{1}{l}{\multirow{1}*{Scene\_no}} &
		\multicolumn{1}{c}{\multirow{1}*{LF-SLAM}}&
		\multicolumn{1}{c}{\multirow{1}*{PL-SLAM Mono}}&
		\multicolumn{1}{c}{\multirow{1}*{{ORB-SLAM2$^\dagger$}}}\\
	&	&\cite{Pumarola2017} (Re-imp)& ~\cite{Mur-Artal2017} \\
		\hline
		
		chess\_01 & \textbf{4.20} &4.58 &  5.06 \\
		chess\_02 & 4.21 &\textbf{3.58}&  4.06 \\
		chess\_03 & \textbf{3.66} &7.58&  7.74\\
		chess\_04 & \textbf{5.01} &6.53&7.25\\
		\hline
		
		fire\_01 & \textbf{4.12} &4.60&  4.94 \\
		fire\_02 & \textbf{2.46} &3.08&  3.10\\
		fire\_03 & 4.14 & \textbf{3.53} &3.56 \\
		fire\_04 & 4.43 & \textbf{4.42} &  4.44 \\
		\hline
		
		heads\_01 &  5.96 & \textbf{4.46}  & 6.66\\
		heads\_02 &  \textbf{3.66} &4.05  & 4.98 \\
		\hline
		
		office\_01 & \textbf{9.38} &10.10  & 9.74 \\
		office\_02 & \textbf{9.21} &12.69& 11.44 \\
		office\_03 & \textbf{7.95} &7.97  & 11.80 \\
		office\_04 & \textbf{11.71}&10.54 & 16.06 \\
		\hline
		
		stairs\_01 & \textbf{12.61}&19.31  & 48.04 \\
		stairs\_02 & \textbf{9.10}&- & - \\
		stairs\_03 & \textbf{5.06}&- & - \\
		stairs\_04 & \textbf{6.42} &- & - \\
		\hline
		
		pumpkin\_01 & \textbf{9.91}&13.22  &13.25 \\
		pumpkin\_02 & \textbf{2.03}&3.33  & 4.62 \\
		pumpkin\_03 & \textbf{8.03}&10.14 & 10.21\\
		pumpkin\_06 & \textbf{8.31} &8.40 & 8.90\\
		\hline
		
		redkitchen\_01 & \textbf{3.92}&5.77 & 6.03 \\
		redkitchen\_02 & \textbf{13.04}&13.24 & 15.63 \\
		redkitchen\_03 & \textbf{4.95}&7.01 & 7.53 \\
		redkitchen\_04 & \textbf{3.37}&5.10 & 5.81 \\
		\hline
		average & \textbf{6.41}&7.52 & 9.60 \\
		\hline
		\hline
		\end{tabular}
	\footnotesize{}
	\end{threeparttable}
	\label{table:7-Scenes}
\end{table}

\noindent\textbf{Experiments on the 7-Scenes benchmark.}
We further evaluate our method on the 7-Scenes benchmark~\cite{Sturm2012}. The sequences are captured by a handheld camera in indoor scenes in 7 different indoor environments with diverse sequences for each environment. As shown in Table~\ref{table:7-Scenes}, our LF-SLAM outperforms other methods on 18 out of 21 sequences, and obtains the lowest average error among the three methods.

The \textit{stairs} sequence is quite challenging due to the repetitive structures of the stairs. Both ORB-SLAM2 and  the reimplemented PL-SLAM Mono fail on this sequence. In contrast, the proposed LF-SLAM successfully handles this challenging sequence with line flows.

The \textit{pumpkin} and \textit{redkitchen} sequences both contain sufficient features with regular and constantly visible lines. Hence, the incorporation of point and line features brings better accuracy. The reimplemented PL-SLAM Mono also provides more accurate camera trajectories compared to ORB-SLAM2. In contrast, in the \textit{fire} sequences, stable lines can hardly be observed in the sequences. The LF-SLAM and PL-SLAM Mono systems can only depend on point features, and achieve similar results with ORB-SLAM2.

\setlength{\tabcolsep}{6pt}
\begin{table}[!t]
	\captionsetup{labelsep=newline}
	\captionsetup{justification=centering}
	\centering
	\scriptsize
	\caption{
		\MakeUppercase{The comparison results on KITTI dataset.
			average translational RMSE drifts are expressed in $\%$ and
			average rotational RMSE drifts are expressed in} $deg/100m$.
		}
	\begin{threeparttable}
	\begin{tabular}{*{2}{c}rrrrrr}
		\hline
		\hline
		\multicolumn{2}{c}{\multirow{3}*{Sequence}} &
		\multicolumn{2}{c}{\multirow{2}*{LF-SLAM}}&
		\multicolumn{2}{c}{{PL-SLAM Mono}} &
		\multicolumn{2}{c}{{{ORB-SLAM2 }}}  \\
		& & & & 	\multicolumn{2}{c}{ \cite{Pumarola2017} (Re-imp) } &
		\multicolumn{2}{c}{ \cite{Mur-Artal2017} }  \\
		& & $t_{rel}$ & $r_{rel}$ & $t_{rel}$ & $r_{rel}$ & $t_{rel}$ & $r_{rel}$ \\
		\hline
		\multicolumn{2}{c}{00}&\textbf{4.233}&\textbf{1.097}&4.342&1.120&4.726&1.296\\
		\multicolumn{2}{c}{01}&\textbf{91.732}&\textbf{1.582}& 95.122 &1.601 & 97.329 & 2.075\\
		\multicolumn{2}{c}{02}&\textbf{5.373}&\textbf{0.612}&5.435&0.620&5.836&0.623\\
		\multicolumn{2}{c}{03}&\textbf{2.182}&0.632&2.215&\textbf{0.513}&2.321&0.569\\
		\multicolumn{2}{c}{04}&\textbf{1.523}&\textbf{0.221}&1.618&0.241&1.719&0.245\\
		\multicolumn{2}{c}{05}&\textbf{3.198}&\textbf{0.544}&3.310&0.600&3.312&0.619\\
		\multicolumn{2}{c}{06}&7.058&\textbf{0.590}&\textbf{6.930}&0.651&9.178&0.661\\
		\multicolumn{2}{c}{07}&\textbf{3.282}&1.745&3.358&2.013&3.360&\textbf{1.318}\\
		\multicolumn{2}{c}{08}&\textbf{13.040}&\textbf{0.617}&13.420&0.642&13.479&0.644\\
		\multicolumn{2}{c}{09}&\textbf{4.452}&\textbf{2.595}&5.622&3.012&5.722&3.324\\
		\multicolumn{2}{c}{10}&\textbf{3.701}&\textbf{0.649}&3.752&0.666&3.860&0.666\\
		\hline
		\multicolumn{2}{c}{average}&\textbf{12.707}&\textbf{0.989}&13.193&1.061&13.712&1.094\\
		\hline
		\hline
	\end{tabular}
	\end{threeparttable}
\label{table:KITTIATE}
\end{table}

\noindent\textbf{Experiments on the KITTI benchmark.} We evaluate our method on the KITTI benchmark~\cite{Geiger2012}. The KITTI dataset is captured by a stereo camera (only left images used) mounted in front of the car. The ground truth trajectories of these sequences are given for performance evaluation. We evaluate the translational RMSE drift errors and rotational RMSE drift errors to compare these algorithms in the toolkit~\cite{Geiger2012}.
Our LF-SLAM exhibits better performance than the baselines (shown in Table~\ref{table:KITTIATE}). We cannot find many line segments in these outdoor datasets. Fortunately, certain line flows, including the long-term stable line segments on the roads, improve the performance. Although the improvements are not significant, they still prove the effectiveness of our line flows.

\setlength{\tabcolsep}{11pt}
\begin{table*}[!t]
	\centering
	\small
	\captionsetup{labelsep=newline}
	\captionsetup{justification=centering}
	\caption{\MakeUppercase{The absolute keyframe trajectory error (ATE) evaluation on the EuRoC MAV dataset} (RMSE, m). \MakeUppercase{``-'' denotes that the system fails in the sequence.}}
	\begin{threeparttable}
\begin{tabular}{ccrrrrrrrr}
		\hline
	\hline
	\multicolumn{2}{c}{\multirow{2}[0]{*}{Sequence}} & \multicolumn{1}{c}{ORB-SLAM} & \multicolumn{1}{c}{DSO} & \multicolumn{1}{c}{LDSO} & \multicolumn{1}{c}{DSM} & \multicolumn{1}{c}{PL-SLAM} & \multicolumn{1}{c}{\multirow{2}[0]{*}{LF-P-SLAM}} & \multicolumn{1}{c}{\multirow{2}[0]{*}{LF-PL-SLAM}} & \multicolumn{1}{c}{\multirow{2}[0]{*}{LF-SLAM}} \\
	\multicolumn{2}{c}{} &  \multicolumn{1}{c}{\cite{Mur-Artal2017}} &\multicolumn{1}{c}{\cite{Engel2018}}   & \multicolumn{1}{c}{\cite{XiangGao2018}}   & \multicolumn{1}{c}{\cite{Zubizarreta2020}}  &
	\multicolumn{1}{c}{\cite{Pumarola2017}(Re-imp)}   &  & &  \\
	\hline
	\multicolumn{2}{c}{MH-01-easy} & 0.070 & 0.046 & 0.053 & \textbf{0.039} & 0.046 & 0.045 & 0.041 & 0.044 \\
	\multicolumn{2}{c}{MH-02-easy} & 0.066 & 0.046 & 0.062 & 0.036 & 0.036 & 0.035 & 0.042 & \textbf{0.034} \\
	\multicolumn{2}{c}{MH-03-med} & 0.071 & 0.172 & 0.114 & 0.055 & 0.042 & 0.049 & 0.043 & \textbf{0.039} \\
	\multicolumn{2}{c}{MH-04-diff} & 0.081 & 3.810 & 0.152 & 0.057 & 0.071 & 0.064 & 0.099 & \textbf{0.055} \\
	\multicolumn{2}{c}{MH-05-diff} & 0.060 & 0.110 & 0.085 & 0.067 & 0.084 & 0.092 & 0.061 & \textbf{0.051} \\
	\multicolumn{2}{c}{	V1-01-easy} & \textbf{0.015} & 0.089 & 0.099 & 0.095 & 0.097 & 0.096 & 0.088 & 0.094 \\
	\multicolumn{2}{c}{	V1-02-med} & \textbf{0.020} & 0.107 & 0.087 & 0.059 & 0.064 & 0.064 & 0.063 & 0.063 \\
	\multicolumn{2}{c}{	V1-03-diff} & - & 0.903 & 0.536 & \textbf{0.076} & - & - & - & 0.090 \\
	\multicolumn{2}{c}{	V2-01-easy} & \textbf{0.015} & 0.044 & 0.066 & 0.056 & 0.057 & 0.060 & 0.067 & 0.059 \\
	\multicolumn{2}{c}{	V2-02-med} & \textbf{0.017} & 0.132 & 0.078 & 0.057 & 0.060 & 0.057 & 0.063 & 0.056 \\
	\multicolumn{2}{c}{	V2-03-diff} & - & 1.152 & - & 0.784 & - & - & 1.562 & \textbf{0.293} \\
		\hline
	\hline
\end{tabular}%

\footnotesize{}
\end{threeparttable}
	\label{table:EuRoCMAV}
\end{table*}

\begin{figure}[t]
	\centering
	\begin{minipage}[c]{\linewidth}
		\centering
		\includegraphics[width=\linewidth]{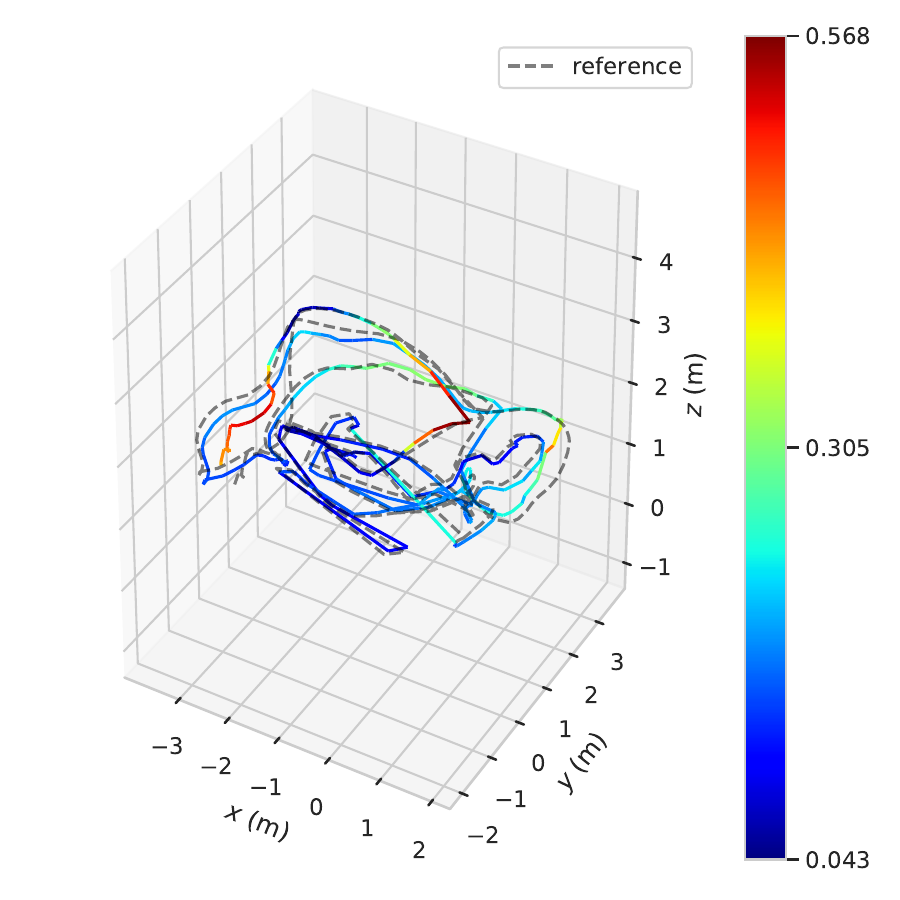}
	\end{minipage}
	\caption{Trajectory obtained by LF-SLAM on \textit{V2\_03\_difficult} and the ground truth as reference. The pseudo colors indicate the absolute pose errors along the LF-SLAM trajectory.}
	\label{fig:EuRoCFig}
\end{figure}

\noindent\textbf{Experiments on the EuRoC MAV dataset.} The EuRoC MAV dataset consists of 11 stereo-inertial sequences recorded in different indoor environments with structural information. These sequences were captured by randomly walking in the rooms. Therefore, loops with different sizes are recorded in these sequences. Table~\ref{table:EuRoCMAV} shows the RMSEs of the camera trajectory on the sequences with different motions on the left images.
DSO, LDSO~\cite{XiangGao2018}, and DSM~\cite{Zubizarreta2020} are direct methods. ORB-SLAM, PL-SLAM Mono, and LF-SLAM are indirect methods.  The results of LDSO, DSO and DSM are obtained using a sequential  implementation without enforcing real-time operation, which means these methods are run with a single thread. ORB-SLAM, PL-SLAM Mono and LF-SLAM are performed with multiple-thread.
DSM successfully runs all sequences and achieves the best performances among the direct methods because of the beneficial strategy of reusing long-term and stable features~\cite{Zubizarreta2020}.
ORB-SLAM achieves the best performances on 4 of the 11 sequences, but fails on 2 sequences with long-term fast and abrupt motions.
PL-SLAM performs better than ORB-SLAM on 4 sequences due to the additional line features.
LF-SLAM achieves the best performances on 5 of the 11 sequences. Compared with ORB-SLAM and PL-SLAM, LF-SLAM is much more robust since it successfully runs all sequences. We visualize the trajectory of the \textit{V2\_03\_diff}, the most challenging sequence that fails most of the other methods, obtained by LF-SLAM along with the ground truth in Fig.~\ref{fig:EuRoCFig}. It can be seen that the estimated trajectory constantly adheres to the ground truth. Note that both DSM and LF-SLAM can successfully run all sequences as seen from Table~\ref{table:EuRoCMAV}. However, LF-SLAM gains on 7 of the 11 sequences. Moreover, on sequences with challenging camera motions, e.g., \textit{V2\_03\_diff}, LF-SLAM obtains much lower errors than DSM.

\begin{figure}[t]
	\centering
	\begin{minipage}[c]{\linewidth}
		\centering
		\includegraphics[width=\linewidth]{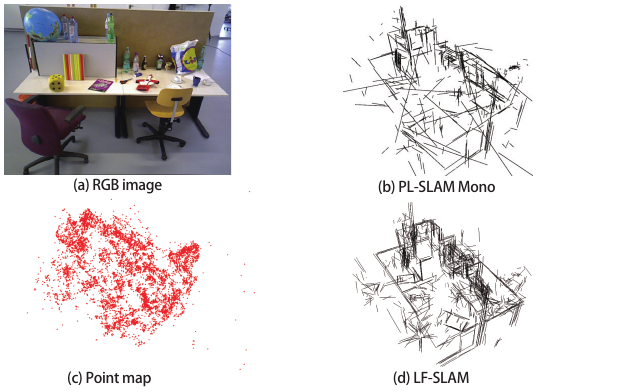}
	\end{minipage}
	\caption{
		The on-the-fly reconstruction result on \textit{fr3\_long\_office\_household} sequence.
	 (a) One typical view of the scene.
	 (b) The line map of PL-SLAM Mono. (c) The point map of LF-SLAM. (d) The line map of LF-SLAM.
	}
	\label{fig:LongOfficeHousehold}
\end{figure}

\begin{figure}[t]
	\centering
	\begin{minipage}[c]{\linewidth}
		\centering
		\includegraphics[width=\linewidth]{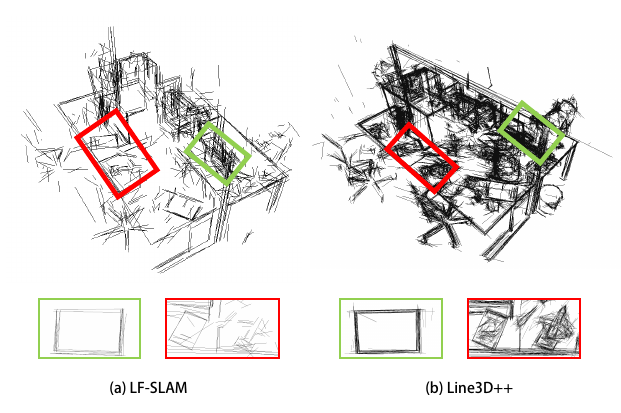}
	\end{minipage}
	\caption{Mapping results of LF-SLAM and Line3D++~\cite{Hofer2017} on the \emph{long\_office\_household} dataset. The results demonstrate that our map is more concise than that of Line3D++.}
	\label{fig:ExperimentComparison}
\end{figure}

\begin{figure*}[t]
	\centering
	\subfloat{
		\centering
		\includegraphics[width=.23\textwidth]{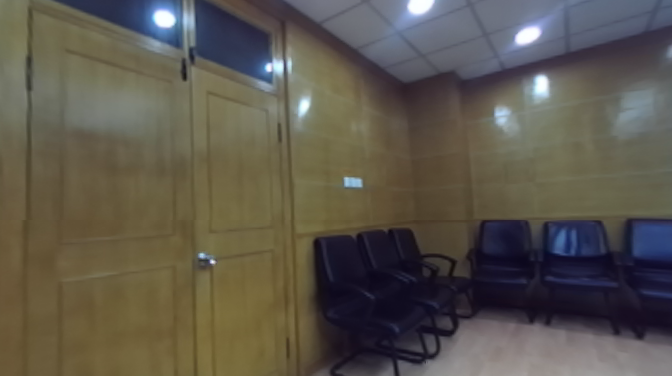}
	}
	\subfloat{
		\centering
		\includegraphics[width=.23\textwidth]{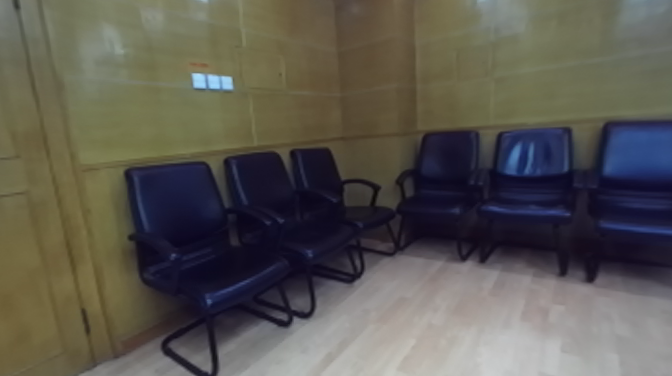}
	}
	\subfloat{
		\centering
		\includegraphics[width=.23\textwidth]{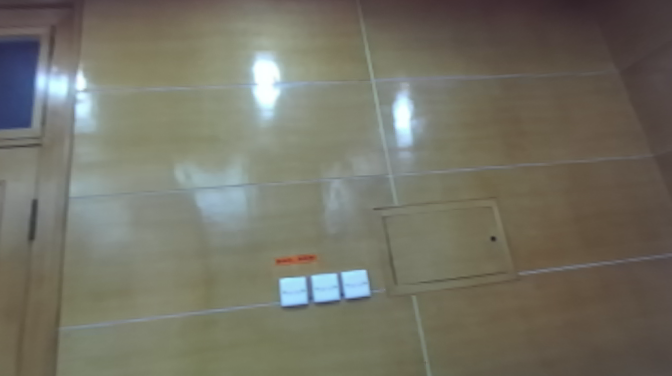}
	}
	\subfloat{
		\centering
		\includegraphics[width=.23\textwidth]{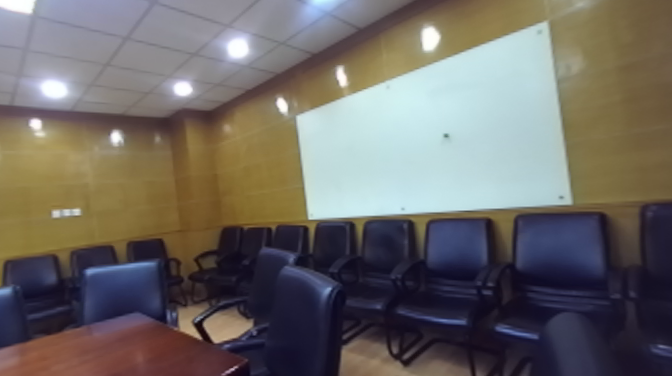}
	}

	\subfloat{
		\centering
		\includegraphics[width=.23\textwidth]{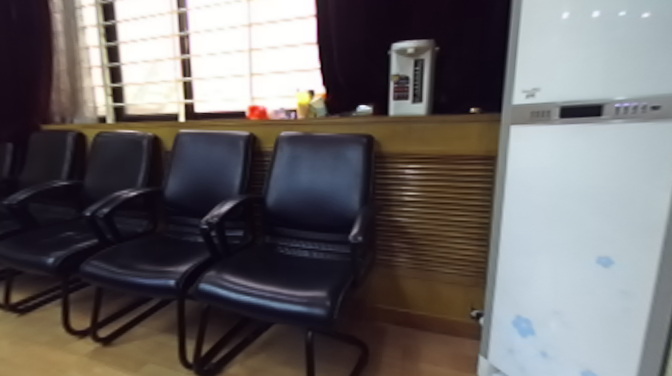}
	}
	\subfloat{
		\centering
		\includegraphics[width=.23\textwidth]{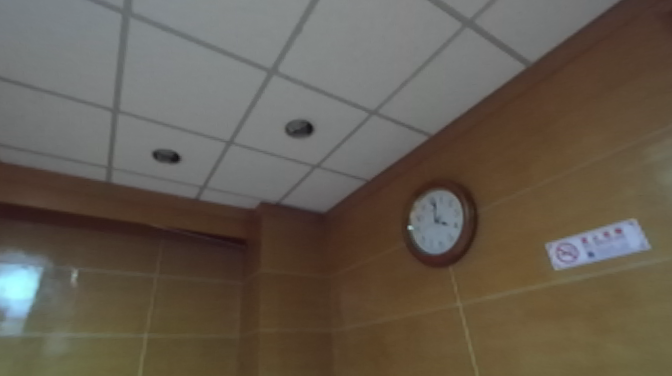}
	}
	\subfloat{
		\centering
		\includegraphics[width=.23\textwidth]{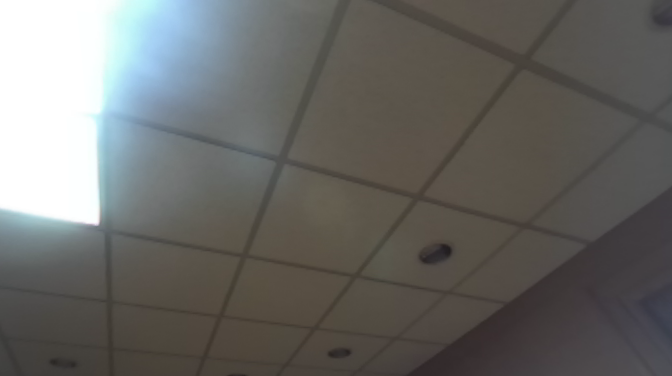}
	}
	\subfloat{
		\centering
		\includegraphics[width=.23\textwidth]{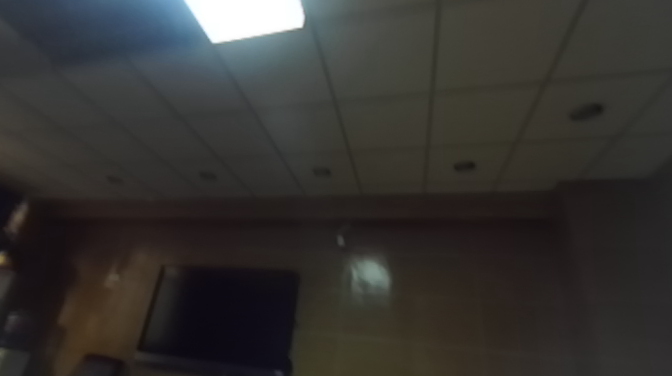}
	}
	\caption{Some images we captured in an office room.  These images contains challenges such as texture-less regions, image blur, illumination variations, similar appearance and overexposure regions.}
	\label{fig:OurDataset2135}
\end{figure*}

\subsection{Qualitative Evaluation in Mapping}
\add{We visualize the 3D maps reconstructed from the \textit{fr3\_long\_office} sequence in Fig.~\ref{fig:LongOfficeHousehold}. Compared to the 3D point map, the 3D line maps contain richer structural information of the scene. Therefore, they are more visually meaningful and can provide more constrains during the dual optimization of camera poses and maps. Compared to PL-SLAM Mono~\cite{Pumarola2017}, LF-SLAM generates a more complete and neater line map.}

We qualitatively compare our method against the state-of-the-art Line3D++ algorithm \cite{Hofer2017}. All the parameters of line3D++ are set by default. Considering that Line3D++ requires an SfM algorithm to generate camera poses and find visual neighbours through sparse 3D points, we adopt the state-of-the-art offline ColMap software \cite{Schonberger2016} for the initialization.

\begin{figure}[t]
	\subfloat[Door]{
	\centering
	\includegraphics[width=.23\textwidth]{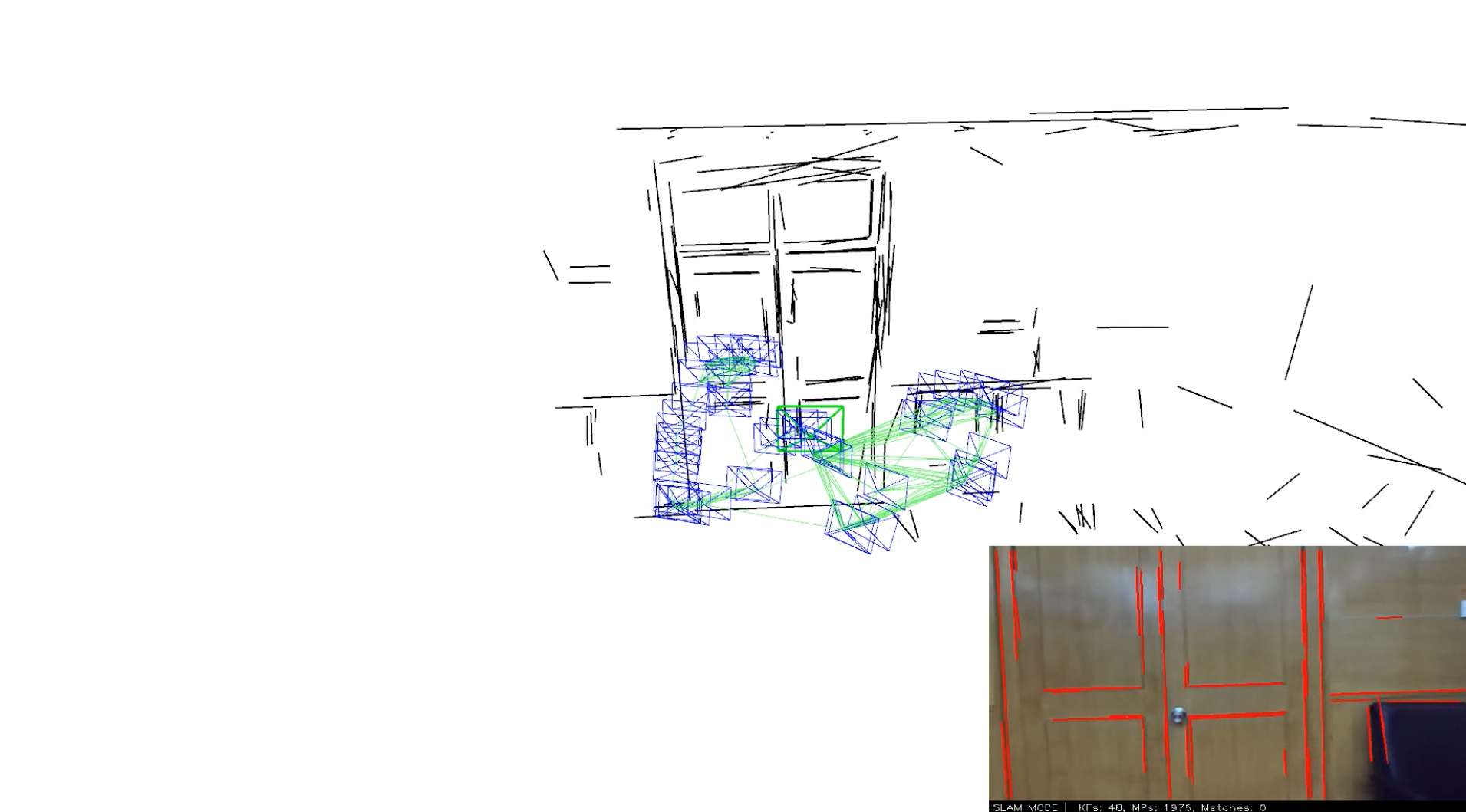}
	}
	\subfloat[Wall]{
		\centering
		\includegraphics[width=.23\textwidth]{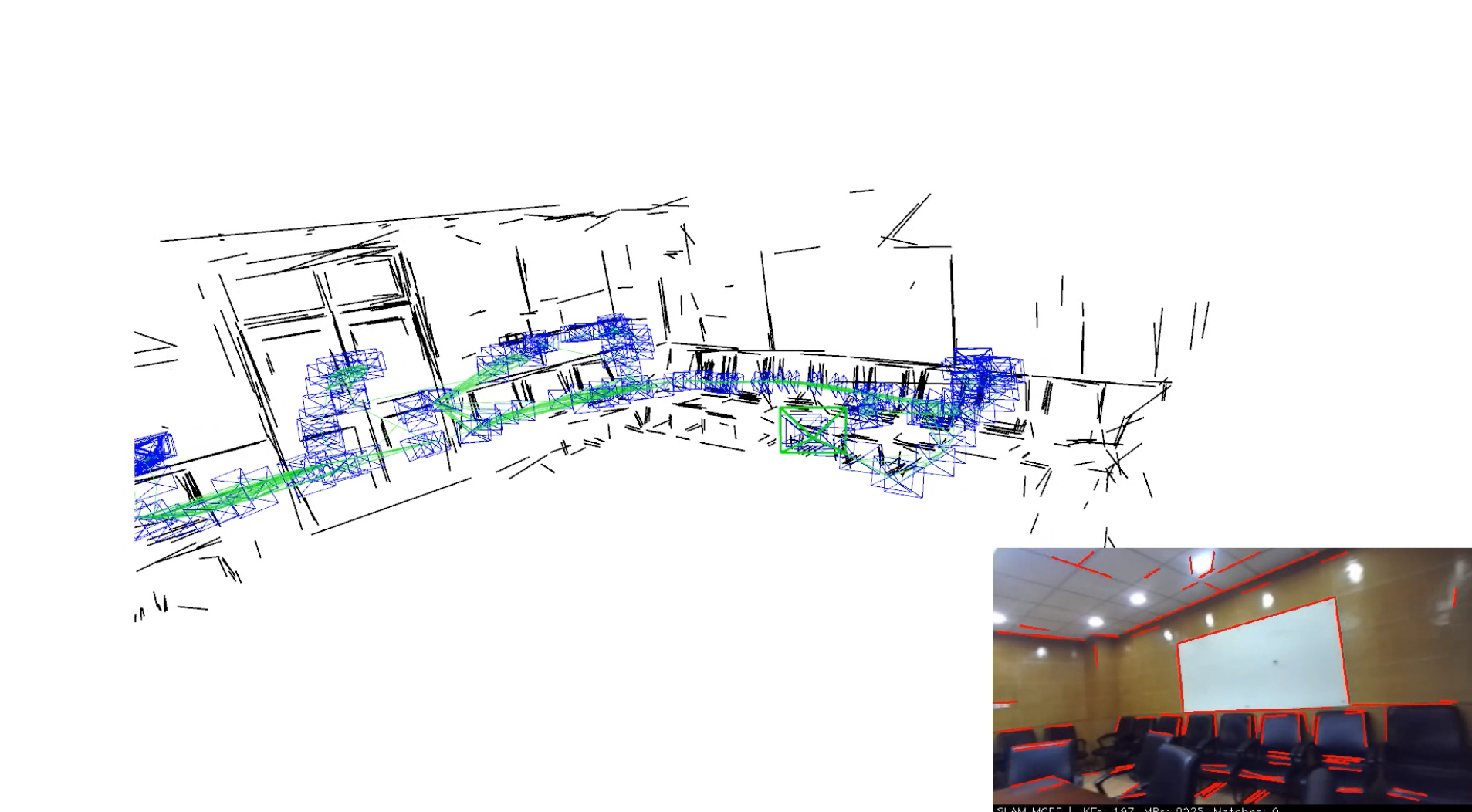}
	}

	\subfloat[Ceiling]{
	\centering
	\includegraphics[width=.23\textwidth]{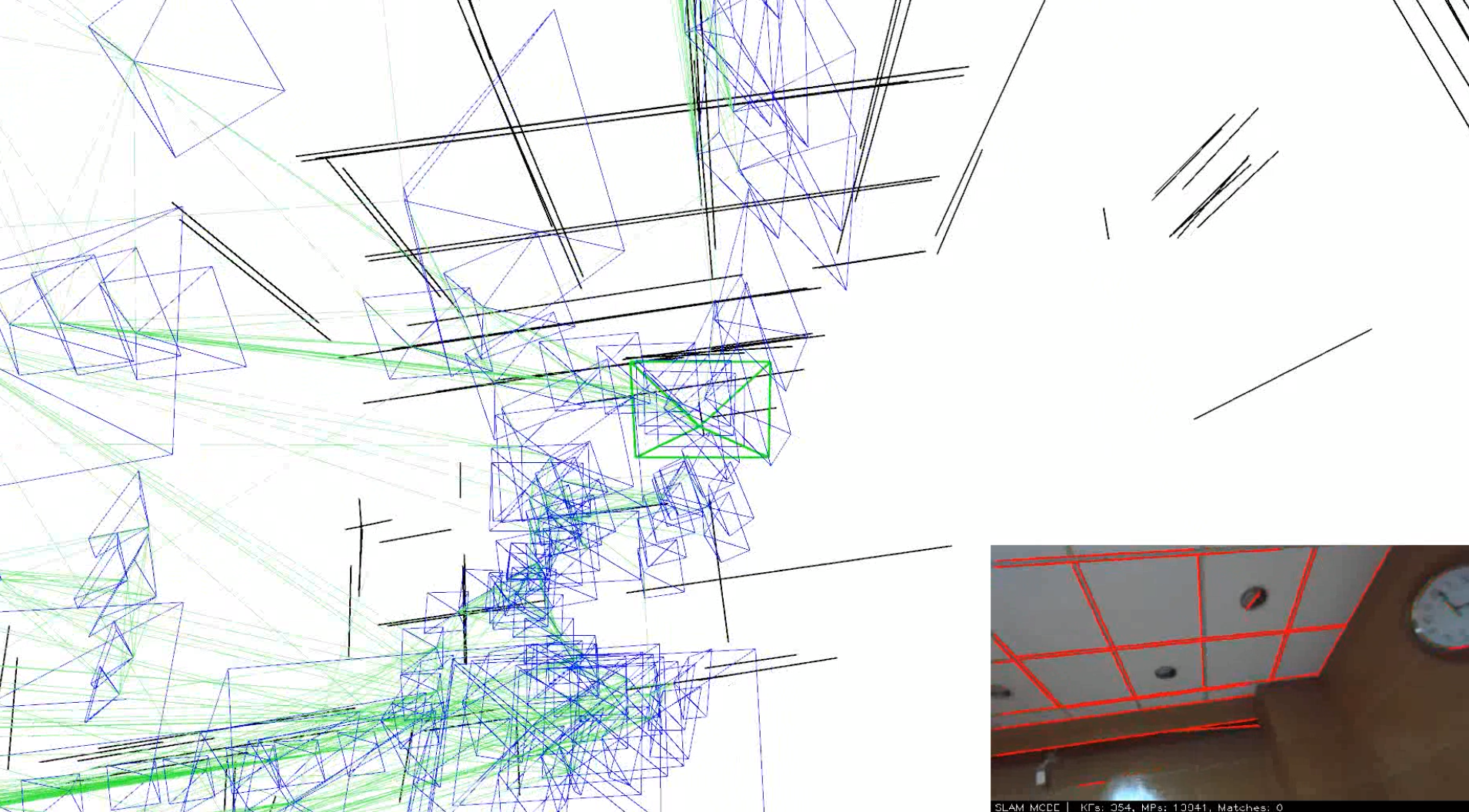}
	}
	\subfloat[Ceiling]{
		\centering
		\includegraphics[width=.23\textwidth]{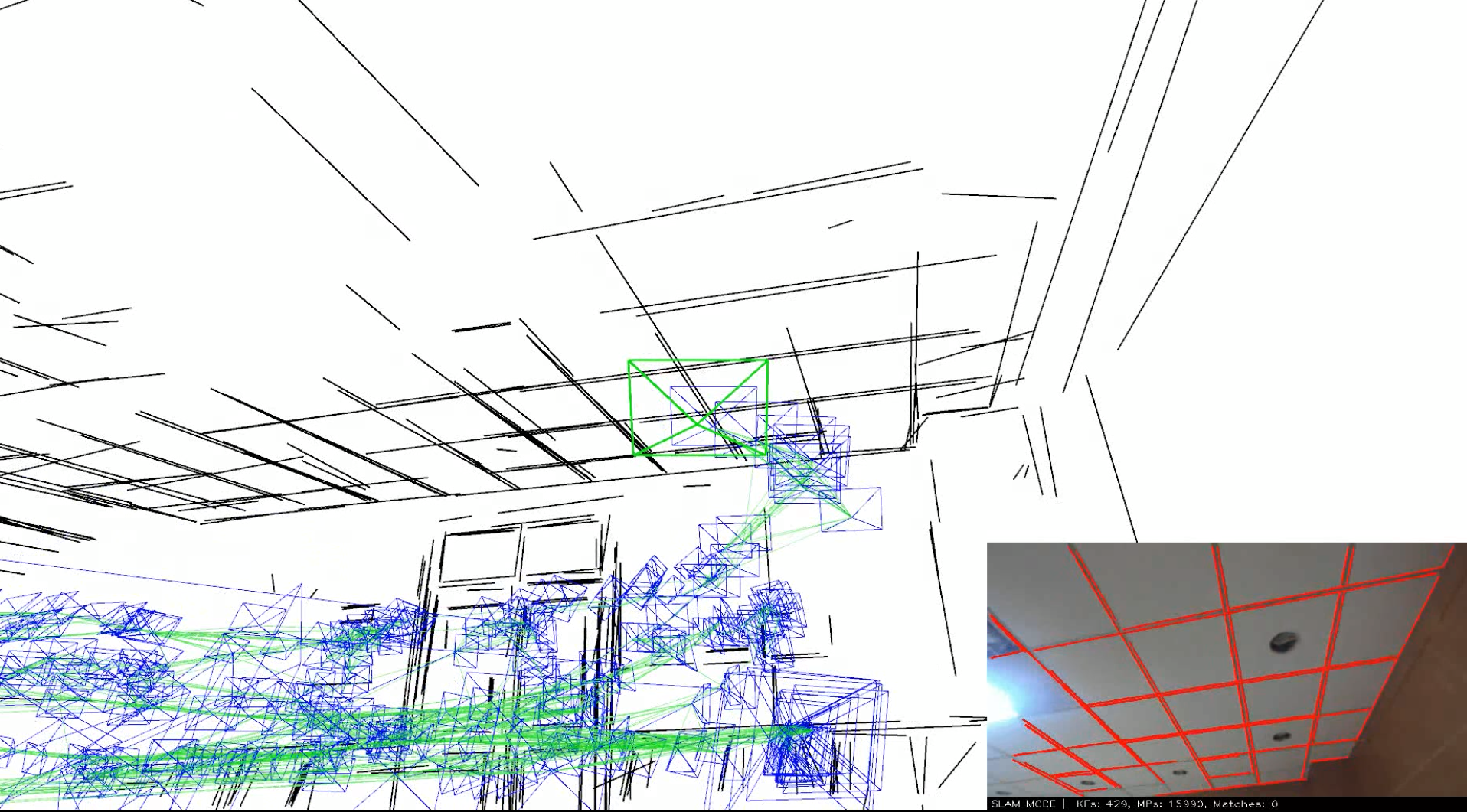}
	}
	\caption{The incremental reconstruction. The current images are shown on bottom-right, while 3D line map and camera trajectory are shown on top-left. (a) Door recovery; (b) Wall Recovery; (c-d) Ceiling recovery.}
	\label{fig:OurDataset2135Result}
\end{figure}

First, we put all the images into LF-SLAM and Line3D++ to generate a 3D line map, as illustrated in Fig.~\ref{fig:ExperimentComparison}.  Although both 3D line maps recover most objects, such as the desk, books, and box, our map is more concise than the map of Line3D++. LF-SLAM extracts  824 3D line segments, while Line3D++ brings 6721 3D line segments. We zoom in some details on the desk with the corresponding green and red rectangles. Although Line3D++ builds upon a line map with very accurate poses,  the generated line map has more redundant 3D line segments because Line3D++ does not consider the line correspondences across successive frames.

To reconstruct a scene, we collect an RGB image sequence in an office room with a monocular camera (Kinect V2). Fig.~\ref{fig:OurDataset2135} demonstrates a few images captured in the rooms.
Fig.~\ref{fig:OurDataset2135Result} visualizes the on-the-fly reconstruction of our algorithm. The red line segments on the bottom right image are the extracted line segments, while the dashed lines are the reconstructed 3D line segments.
Note that although a few line segments are not extracted in a single frame and some line segments are incomplete, such line segments are refined based on the information in multiple frames.
In Fig.~\ref{fig:OurDataset2135Result} (a) to (d), we extract 2D line segments from the images successfully despite challenges such as similar textures, illumination variations, and reflected light.

\begin{figure}[t]
	\subfloat[Line map with trajtory]{
	\centering
	\includegraphics[width=.23\textwidth]{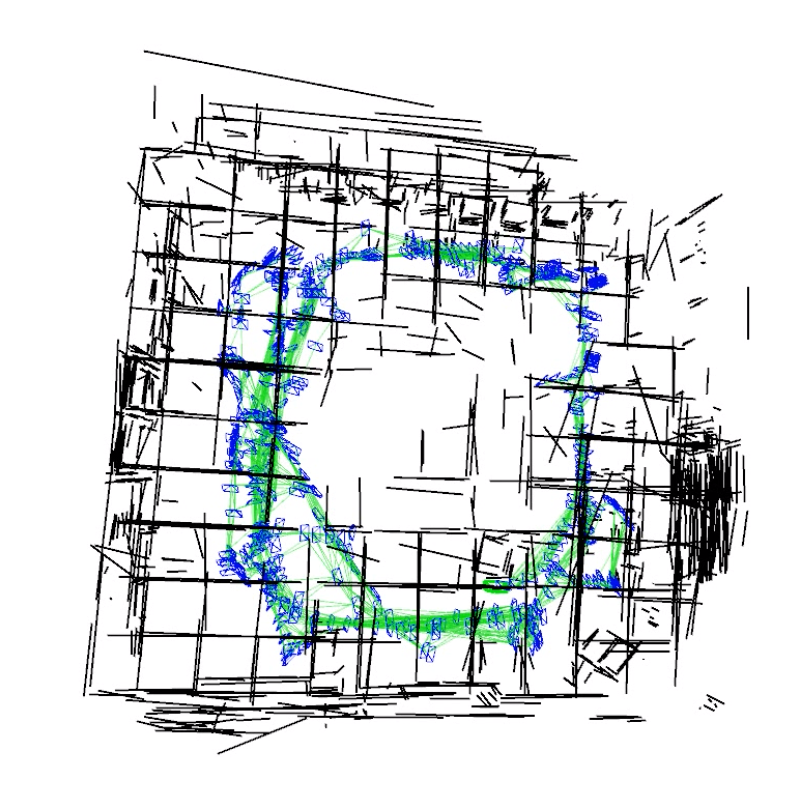}
	}
	\subfloat[Top view]{
		\centering
		\includegraphics[width=.23\textwidth]{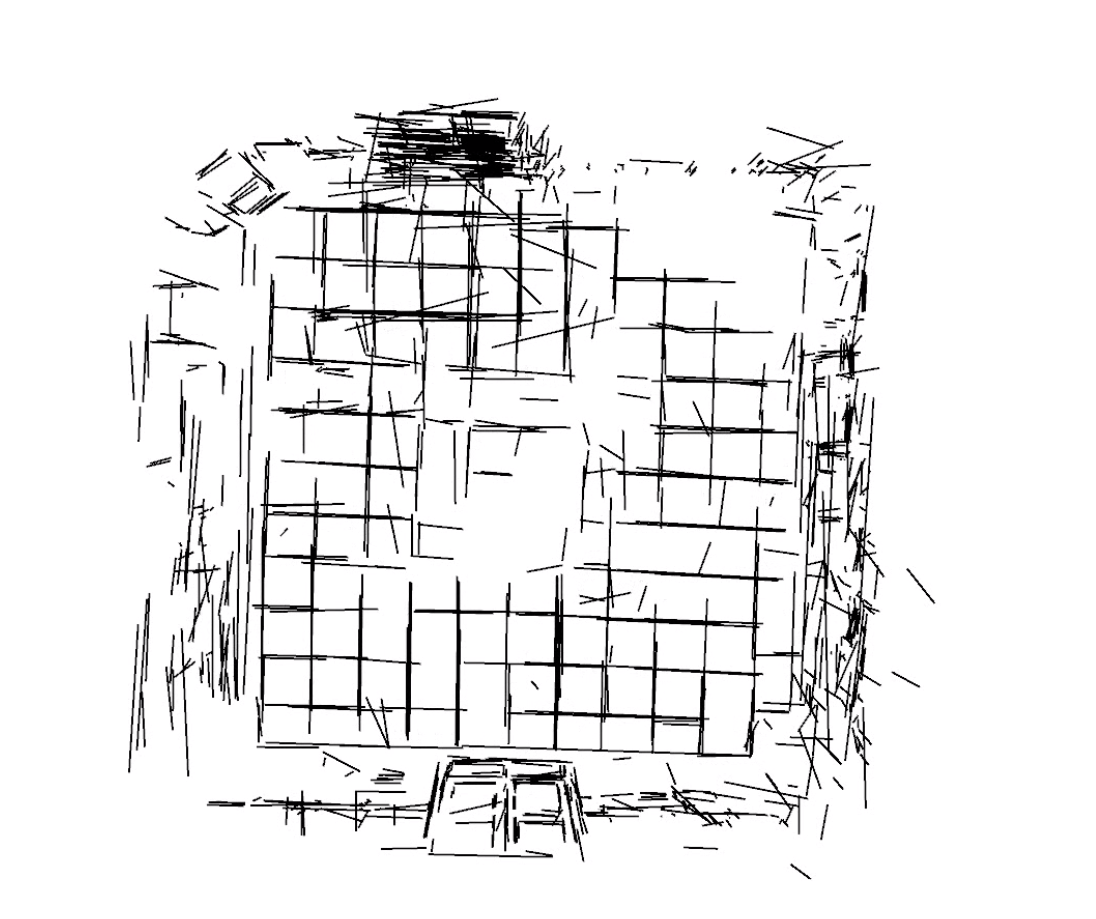}
	}

	\subfloat[Left view]{
	\centering
	\includegraphics[width=.23\textwidth]{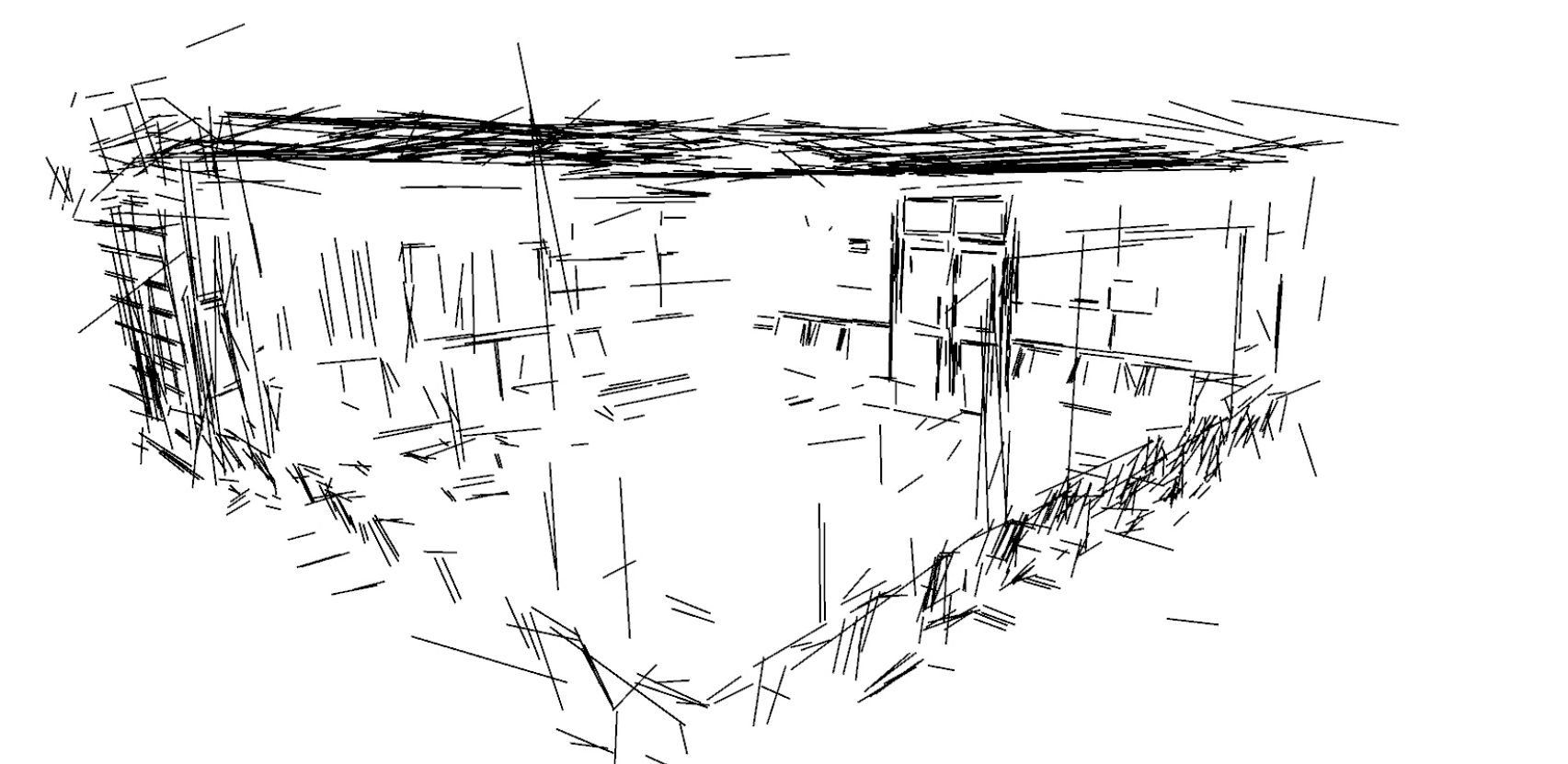}
	}
	\subfloat[Right view]{
		\centering
		\includegraphics[width=.23\textwidth]{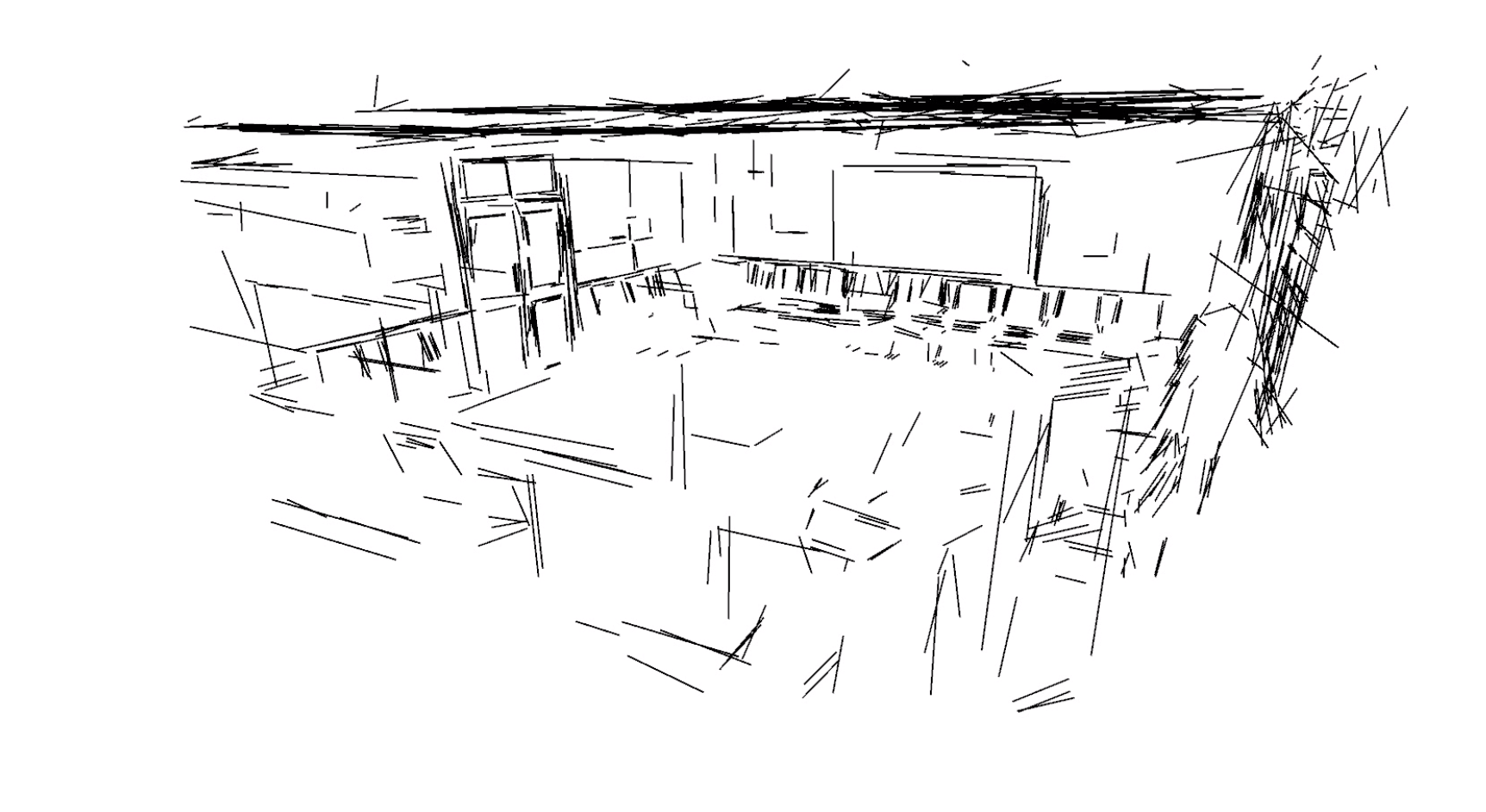}
	}
	\caption{ Reconstructed 3D line map. (a) The 3D line map and the camera trajectory. (b-d) The different views of the 3D line map.}
	\label{fig:OurDataset2135Map}
\end{figure}

We visualize the final 3D line map in Fig.~\ref{fig:OurDataset2135Map}. We recover most of the parts of the scenes in the office, such as, the ceiling, desk, and white board. It also validates the good reconstruction ability of our method due to the reliable correspondences in the line flows.

All these qualitative results verify our contribution to fully exploiting the spatial-temporal coherence of line features to maintain stable and reliable line flows, which is the key to accurate pose estimation and map reconstruction.

\section{Conclusions}
\label{Sec:Conclusion}
We propose a novel line flow representation for describing line segments in consecutive frames.
Line flows encode the spatial-temporal coherence of line segments in image sequences by considering the correspondences among 2D and 3D line segments. Based on line flows, we developed LF-SLAM, which can address many challenging scenarios, such as texture-less regions, occlusions, blurred images, and repetitive appearances. Experiments show that compared with other state-of-the-art direct and indirect methods, LF-SLAM has higher efficiency and localization accuracy. In addition, LF-SLAM generates  precise and visually appealing 3D line maps on-the-fly.

The proposed LF-SLAM still has limitations. Line flows maintain 2D line segments by exploiting the spatial-temporal constraints. Therefore, our system fails when the camera motion breaks the constraints, especially when long-term abrupt camera motion occurs. One possible solution to this problem is adopting a coarse-to-fine strategy by extracting line segments from image pyramids. \add{On the other hand, our line flows model the coherence of line segments in monocular sequences, which cannot obtain camera trajectories and 3D maps with real metric scale.} Hence, in the future, we will extend the representation to SLAM systems with stereo and RGBD cameras. Moreover, we plan to compositely model flows of rich types of features, e.g., lines and planes. These are expected to enhance the robustness of SLAM systems to large camera motion and their ability to generate more complete 3D maps.

\appendices
\section{Non-linear Optimizations}

\subsection{Line Representation for Optimization}
We use an iterative method to efficiently minimize a non-linear energy function.
The Pl{\"{u}}cker coordinates has 6 DOF variables with 2 strict constraints. In our optimization process, it is difficult for the iterative method to keep the strict constraints. To deal with this problem, we transform line representation from Pl{\"{u}}cker coordinates into orthonormal representation which has 4 DOF variables.
We rearrange 3D line $\mathbf{L} = [ \mathbf{n} \vert \mathbf{v}] $ and leverage QR decomposition to obtain orthonormal representation:
\begin{equation}
\begin{aligned}
	\mathbf{L}
	&= \sqrt{\left\|\mathbf{n}\right\|^2 + \left\|\mathbf{v}\right\|^2} \mathbf{U} \begin{bmatrix}
	 \sigma_1 & \\ & \sigma_2\\
	\end{bmatrix} \\
	&
	 \begin{aligned}
	=&\sqrt{\left\|\mathbf{n}\right\|^2 + \left\|\mathbf{v}\right\|^2}  \\ &\begin{bmatrix}
	\frac{\mathbf{n}}{\left\|\mathbf{n}\right\|} &
	\frac{\mathbf{v}}{\left\|\mathbf{v}\right\|} &
	\frac{\mathbf{n} \times \mathbf{v}}{\left\|\mathbf{n} \times \mathbf{v}\right\|}
	\end{bmatrix}
	\begin{bmatrix}
	 \frac{\left\|\mathbf{n}\right\|}{\sqrt{\left\|\mathbf{n}\right\|^2 + \left\|\mathbf{v}\right\|^2}} & \\
	 & \frac{\left\|\mathbf{v}\right\|}{\sqrt{\left\|\mathbf{n}\right\|^2 + \left\|\mathbf{v}\right\|^2}} \\
	\end{bmatrix},
	\end{aligned}
	\\
	\mathbf{W} &= \begin{bmatrix}
	 \sigma_1  & -\sigma_2 \\ \sigma_2 & \sigma_1
	\end{bmatrix},
\end{aligned}
\end{equation}
where, $\mathbf{U} \in \mathrm{SO(3)}$ and $\mathbf{W} \in \mathrm{SO(2)}$. The orthonormal representation $(\mathbf{U},\mathbf{W}) \in \mathrm{SO(3)} \times \mathrm{SO(2)}$ corresponds to a unique 3D line. We refer reader to \cite{Bartoli2005} for more details.

\subsection{Jacobian Matrices of Error Terms}
We deduce point and line error terms of Jacobian matrices using left version definition~\cite{Sol}. We utilize the Lie algebra $\mathfrak{se}(3)$ of the corresponding rigid transformation $\mathbf{T}$ to minimize the energy function. We derived point error term $\bm{\epsilon}_p $ with respect to 3D point $\mathbf{J_p^{\epsilon_p}}$ and camera pose $\mathbf{J_T^{\epsilon_p}}$, and line error term $\bm{\epsilon}_l $ with respect to 3D line $\mathbf{J_l^{\epsilon_l}}$ and camera pose $\mathbf{J_T^{\epsilon_l}}$. We have:
\begin{equation}
\begin{aligned}
\mathbf{J_p^{\epsilon_p}} &= \mathbf{J_p^\circ} \mathbf{R}, \\
\mathbf{J_T^{\epsilon_p}} &= \mathbf{J_p^\circ} \begin{bmatrix} [-\mathbf{RP}]_\times \vert \mathbf{I}
\end{bmatrix},\\
\mathbf{J_L^{\epsilon_l}} &= \mathbf{J_L^\circ} \begin{bmatrix} \mathbf{R} \vert [\mathbf{t}]_\times \mathbf{R} \end{bmatrix} \begin{bmatrix}
\mathbf{0}_{3\times 1} & -\sigma_1 \mathbf{u}_3 & \sigma_1 \mathbf{u}_2 & -\sigma_2 \mathbf{u}_1 \\
\sigma_2 \mathbf{u}_3 & \mathbf{0}_{3 \times 1} & -\sigma_2 \mathbf{u}_1 & \sigma_1 \mathbf{u}_2 \\
\end{bmatrix}, \\
\mathbf{J_T^{\epsilon_l}} &= \mathbf{J_L^\circ} \begin{bmatrix} - [\mathbf{Rn}]_\times - [\mathbf{t}]_\times [\mathbf{Rv}]_\times \vert -[\mathbf{Rv}]_\times \end{bmatrix}, \\
\end{aligned}
\end{equation}
where, $\mathbf{u}_i$ is the $i$-th column of $\mathbf{U}$,
\begin{equation}
\begin{aligned}
	\mathbf{J_p^\circ} &= \begin{bmatrix}
	\frac{1}{p_z} & 0  & \frac{-p_x}{p_z^2} \\
	0  & \frac{1}{p_z} & \frac{-p_y}{p_z^2} \\
	\end{bmatrix} \mathbf{K}_p, \\
	\mathbf{J_L^\circ} &= (l_x^2+l_y^2)^{-3/2} \begin{bmatrix} \mathbf{e}_1 \vert\mathbf{e_2}	\end{bmatrix}^T \begin{bmatrix}
	l_y^2  & -l_xl_y & 0 \\
	-l_xl_y & l_x^2  & 0 \\
	-l_xl_z & -l_yl_z & l_x^2+l_y^2 \\
	\end{bmatrix} \mathbf{K}_l,
\end{aligned}
\end{equation}
where, $(p_x,p_y,p_z)$ and $(l_x,l_y,l_z)$ are the homogeneous coordinate of projected 3D point and 3D line, respectively.

%
%
%
%

\ifCLASSOPTIONcaptionsoff
  \newpage
\fi



%

{
	\bibliographystyle{IEEEtran}
	\bibliography{egbib}

\begin{thebibliography}{10}
\providecommand{\url}[1]{#1}
\csname url@samestyle\endcsname
\providecommand{\newblock}{\relax}
\providecommand{\bibinfo}[2]{#2}
\providecommand{\BIBentrySTDinterwordspacing}{\spaceskip=0pt\relax}
\providecommand{\BIBentryALTinterwordstretchfactor}{4}
\providecommand{\BIBentryALTinterwordspacing}{\spaceskip=\fontdimen2\font plus
\BIBentryALTinterwordstretchfactor\fontdimen3\font minus
  \fontdimen4\font\relax}
\providecommand{\BIBforeignlanguage}[2]{{%
\expandafter\ifx\csname l@#1\endcsname\relax
\typeout{** WARNING: IEEEtran.bst: No hyphenation pattern has been}%
\typeout{** loaded for the language `#1'. Using the pattern for}%
\typeout{** the default language instead.}%
\else
\language=\csname l@#1\endcsname
\fi
#2}}
\providecommand{\BIBdecl}{\relax}
\BIBdecl

\bibitem{Engel2018}
J.~Engel, V.~Koltun, and D.~Cremers, ``{Direct Sparse Odometry},'' \emph{{IEEE}
  Transactions on Pattern Analysis and Machine Intelligence (T-PAMI)}, vol.~40,
  no.~3, pp. 611--625, 2018.

\bibitem{Engel2014}
J.~Engel, T.~Sch{\"{o}}ps, and D.~Cremers, ``{LSD-SLAM: Large-scale Direct
  Monocular SLAM},'' in \emph{Proceedings of the European Conference on
  Computer Vision (ECCV)}, 2014, pp. 834--849.

\bibitem{Newcombe2011}
R.~A. Newcombe, S.~J. Lovegrove, and A.~J. Davison, ``{DTAM: Dense Tracking and
  Mapping in Real-time},'' in \emph{Proceedings of the International Conference
  on Computer Vision (ICCV)}, 2011, pp. 2320--2327.

\bibitem{Mur-Artal2017}
R.~Mur-Artal and J.~D. Tard\'os, ``{ORB-SLAM2: An Open-source SLAM System for
  Monocular, Stereo, and RGB-D Cameras},'' \emph{{IEEE} Transactions on
  Robotics (T-RO)}, vol.~33, no.~5, pp. 1255--1262, 2017.

\bibitem{VonGioi2010}
R.~G. von Gioi, J.~Jakubowicz, J.-M. Morel, and G.~Randall, ``{LSD: A Fast Line
  Segment Detector with a False Detection Control},'' \emph{{IEEE} Transactions
  on Pattern Analysis and Machine Intelligence (T-PAMI)}, vol.~32, no.~4, pp.
  722--732, 2010.

\bibitem{Zhang2013}
L.~Zhang and R.~Koch, ``{An Efficient and Robust Line Segment Matching Approach
  based on LBD Descriptor and Pairwise Geometric Consistency},'' \emph{Journal
  of Visual Communication and Image Representation (VCIR)}, vol.~24, no.~7, pp.
  794--805, 2013.

\bibitem{Almazan2017}
E.~J. Almazan, R.~Tal, Y.~Qian, and J.~H. Elder, ``{MCMLSD: A Dynamic
  Programming Approach to Line Segment Detection},'' in \emph{Proceedings of
  the {IEEE} Conference on Computer Vision and Pattern Recognition (CVPR)},
  2017, pp. 5854--5862.

\bibitem{Cho2018}
N.~Cho, A.~Yuille, and S.~Lee, ``{A Novel Linelet-based Representation for Line
  Segment Detection},'' \emph{{IEEE} Transactions on Pattern Analysis and
  Machine Intelligence (T-PAMI)}, vol.~40, no.~5, pp. 1195--1208, 2018.

\bibitem{Pumarola2017}
A.~Pumarola, A.~Vakhitov, A.~Agudo, A.~Sanfeliu, and F.~Moreno-Noguer,
  ``{PL-SLAM: Real-time Monocular Visual SLAM with Points and Lines},'' in
  \emph{Proceedings of the {IEEE} International Conference on Robotics and
  Automation (ICRA)}, 2017, pp. 4503--4508.

\bibitem{Zhao2018_ECCV}
Y.~Zhao and P.~A. Vela, ``{Good Line Cutting: Towards Accurate Pose Tracking of
  Line-assisted VO/VSLAM},'' in \emph{Proceedings of the European Conference on
  Computer Vision (ECCV)}, 2018, pp. 527--543.

\bibitem{Gomez-Ojeda2017}
R.~Gomez-Ojeda, F.-A. Moreno, D.~Scaramuzza, and J.~Gonzalez-Jimenez,
  ``{PL-SLAM: A Stereo SLAM System through the Combination of Points and Line
  Segments},'' \emph{{IEEE} Transactions on Robotics (T-RO)}, vol.~35, no.~3,
  pp. 734--746, 2017.

\bibitem{Ballard1981}
D.~H. Ballard, ``{Generalizing the Hough Transform to Detect Arbitrary
  Shapes},'' \emph{Pattern Recognition (PR)}, vol.~13, no.~2, pp. 111--122,
  1981.

\bibitem{Xue_2019_CVPR}
N.~Xue, S.~Bai, F.~Wang, G.-S. Xia, T.~Wu, and L.~Zhang, ``{Learning Attraction
  Field Representation for Robust Line Segment Detection},'' in
  \emph{Proceedings of the {IEEE} Conference on Computer Vision and Pattern
  Recognition (CVPR)}, 2019, pp. 1595--1603.

\bibitem{Zhang_2019_CVPR}
Z.~Zhang, Z.~Li, N.~Bi, J.~Zheng, J.~Wang, K.~Huang, W.~Luo, Y.~Xu, and S.~Gao,
  ``{PPGNet: Learning Point-Pair Graph for Line Segment Detection},'' in
  \emph{Proceedings of the {IEEE} Conference on Computer Vision and Pattern
  Recognition (CVPR)}, 2019, pp. 7105--7114.

\bibitem{Salaun2016}
Y.~Salaun, R.~Marlet, and P.~Monasse, ``{Multiscale Line Segment Detector for
  Robust and Accurate SfM},'' in \emph{Proceedings of International Conference
  on Pattern Recognition (ICPR)}, 2016, pp. 2000--2005.

\bibitem{Akinlar2011}
C.~Akinlar and C.~Topal, ``{EDLines: A Real-time Line Segment Detector with a
  False Detection Control},'' \emph{Pattern Recognition Letters (PRL)},
  vol.~32, no.~13, pp. 1633--1642, 2011.

\bibitem{Wang2009}
Z.~Wang, H.~Liu, and F.~Wu, ``{HLD: A Robust Descriptor for Line Matching},''
  in \emph{Proceedings of International Conference on Computer-Aided Design and
  Computer Graphics (CADCG)}, 2009, pp. 128--133.

\bibitem{Fan2012}
B.~Fan, F.~Wu, and Z.~Hu, ``{Robust Line Matching through Line-point
  Invariants},'' \emph{Pattern Recognition (PR)}, vol.~45, no.~2, pp. 794--805,
  2012.

\bibitem{Micusik2017}
B.~Micusik and H.~Wildenauer, ``{Structure from Motion with Line Segments under
  Relaxed Endpoint Constraints},'' \emph{International Journal of Computer
  Vision (IJCV)}, vol. 124, no.~1, pp. 65--79, 2017.

\bibitem{Perdices2014}
E.~Perdices, L.~M. L{\'{o}}pez, and J.~M. Ca{\~{n}}as, ``{LineSLAM: Visual Real
  Time Localization using Lines and UKF},'' in \emph{Proceedings of First
  Iberian Robotics Conference - Advances in Robotics}, 2013, pp. 663--678.

\bibitem{Zhou2015}
H.~Zhou, D.~Zou, L.~Pei, R.~Ying, P.~Liu, and W.~Yu, ``{StructSLAM: Visual SLAM
  with Building Structure Lines},'' \emph{{IEEE} Transactions on Vehicular
  Technology (VT)}, vol.~64, no.~4, pp. 1364--1375, 2015.

\bibitem{Li2018}
H.~Li, J.~Yao, J.-c. Bazin, X.~Lu, Y.~Xing, and K.~Liu, ``{A Monocular SLAM
  System Leveraging Structural Regularity in Manhattan World},'' in
  \emph{Proceedings of the {IEEE} International Conference on Robotics and
  Automation (ICRA)}, 2018, pp. 2518--2525.

\bibitem{Sola2012}
J.~Sol{\`{a}}, T.~Vidal-Calleja, J.~Civera, and J.~M.~M. Montiel, ``{Impact of
  Landmark Parametrization on Monocular EKF-SLAM with Points and Lines},''
  \emph{International Journal of Computer Vision (IJCV)}, vol.~97, no.~3, pp.
  339--368, 2012.

\bibitem{Ruifang2018}
D.~Ruifang, V.~Fr{\'{e}}mont, S.~Lacroix, I.~Fantoni, D.~Ruifang,
  V.~Fr{\'{e}}mont, S.~Lacroix, I.~Fantoni, and L.~C. L.-b. Monocu,
  ``{Line-based Monocular Graph SLAM},'' in \emph{Proceedings of {IEEE}
  International Conference on Multisensor Fusion and Integration for
  Intelligent Systems (MFI)}, 2017, pp. 494--500.

\bibitem{Ayache1987}
N.~Ayache and B.~Faverjon, ``{Efficient Registration of Stereo Images by
  Matching Graph Descriptions of Edge Segments},'' \emph{Proceedings of
  International Journal of Computer Vision (IJCV)}, vol.~1, no.~2, pp.
  107--131, 1987.

\bibitem{Bartoli2005}
A.~Bartoli and P.~Sturm, ``{Structure-from-Motion using Lines: Representation,
  Triangulation, and Bundle Adjustment},'' \emph{Computer Vision and Image
  Understanding (CVIU)}, vol. 100, no.~3, pp. 416--441, 2005.

\bibitem{Jain2010}
A.~Jain, C.~Kurz, T.~Thormahlen, and H.~Seidel, ``{Exploiting Global
  Connectivity Constraints for Reconstruction of 3D Line Segments from
  Images},'' in \emph{Proceedings of the {IEEE} Conference on Computer Vision
  and Pattern Recognition (CVPR)}, 2010, pp. 1586--1593.

\bibitem{He2017}
S.~He, X.~Qin, Z.~Zhang, and M.~Jagersand, ``{Incremental 3D Line Segment
  Extraction from Semi-dense SLAM},'' in \emph{Proceedings of International
  Conference on Pattern Recognition (ICPR)}, 2018, pp. 1658--1663.

\bibitem{Hofer2017}
M.~Hofer, M.~Maurer, and H.~Bischof, ``{Efficient 3D Scene Abstraction using
  Line Segments},'' \emph{Computer Vision and Image Understanding (CVIU)}, vol.
  157, pp. 167--178, 2017.

\bibitem{Zhang2014}
L.~Zhang and R.~Koch, ``{Structure and Motion from Line Correspondences:
  Representation, Projection, Initialization and Sparse Bundle Adjustment},''
  \emph{Journal of Visual Communication and Image Representation (VCIR)},
  vol.~25, no.~5, pp. 904--915, 2014.

\bibitem{Zhang2015}
G.~Zhang, J.~H. Lee, J.~Lim, and I.~H. Suh, ``{Building a 3-D Line-based Map
  using Stereo SLAM},'' \emph{{IEEE} Transactions on Robotics (T-RO)}, vol.~31,
  no.~6, pp. 1364--1377, 2015.

\bibitem{Hartley2004}
R.~Hartley and A.~Zisserman, \emph{{Multiple View Geometry in Computer
  Vision}}, 2nd~ed.\hskip 1em plus 0.5em minus 0.4em\relax Cambridge University
  Press, 2004.

\bibitem{Kaess2008}
M.~Kaess, A.~Ranganathan, and F.~Dellaert, ``{iSAM: Incremental Smoothing and
  Mapping},'' \emph{{IEEE} Transactions on Robotics (T-RO)}, vol.~24, no.~6,
  pp. 1365--1378, 2008.

\bibitem{Thrun_2005}
S.~Thrun, W.~Burgard, and D.~Fox, \emph{Probabilistic Robotics}.\hskip 1em plus
  0.5em minus 0.4em\relax The MIT press, 2005.

\bibitem{GromponevonGioi2012}
R.~{Grompone von Gioi}, J.~Jakubowicz, J.-M. Morel, and G.~Randall, ``{LSD: A
  Line Segment Detector},'' \emph{Image Processing On Line (IPOL)}, vol.~2, pp.
  35--55, 2012.

\bibitem{JianboShi1994}
{Jianbo Shi} and Tomasi, ``{Good Features to Track},'' in \emph{Proceedings of
  the {IEEE} Conference on Computer Vision and Pattern Recognition (CVPR)},
  1994, pp. 593--600.

\bibitem{Mur-Artal2014}
R.~Mur-Artal and J.~D. Tardos, ``{Fast Relocalisation and Loop Closing in
  Keyframe-based SLAM},'' in \emph{Proceedings of the {IEEE} International
  Conference on Robotics and Automation (ICRA)}, 2014, pp. 846--853.

\bibitem{Sturm2012}
J.~Sturm, N.~Engelhard, F.~Endres, W.~Burgard, and D.~Cremers, ``{A Benchmark
  for the Evaluation of RGB-D SLAM Systems},'' in \emph{Proceedings of {IEEE}
  International Conference on Intelligent Robots and Systems (IROS)}, 2012, pp.
  573--580.

\bibitem{Shotton2013}
J.~Shotton, B.~Glocker, C.~Zach, S.~Izadi, A.~Criminisi, and A.~Fitzgibbon,
  ``{Scene Coordinate Regression Forests for Camera Relocalization in RGB-D
  Images},'' in \emph{Proceedings of the {IEEE} Conference on Computer Vision
  and Pattern Recognition (CVPR)}, 2013, pp. 2930--2937.

\bibitem{Burri2016}
M.~Burri, J.~Nikolic, P.~Gohl, T.~Schneider, J.~Rehder, S.~Omari, M.~W.
  Achtelik, and R.~Siegwart, ``{The EuRoC Micro Aerial Vehicle Datasets},''
  \emph{International Journal of Robotics Research (IJRR)}, vol.~35, no.~10,
  pp. 1157--1163, 2016.

\bibitem{Geiger2012}
A.~Geiger, P.~Lenz, and R.~Urtasun, ``{Are We Ready for Autonomous Driving? The
  KITTI Vision Benchmark Suite},'' in \emph{Proceedings of the {IEEE}
  Conference on Computer Vision and Pattern Recognition (CVPR)}, 2012, pp.
  3354--3361.

\bibitem{grupp2017evo}
M.~Grupp, ``{evo: Python Package for the Evaluation of Odometry and SLAM},''
  \url{https://github.com/MichaelGrupp/evo}, 2017.

\bibitem{ceres-solver}
S.~Agarwal, K.~Mierle, and Others, ``{Ceres Solver},''
  \url{http://ceres-solver.org}, 2012.

\bibitem{XiangGao2018}
X.~Gao, R.~Wang, N.~Demmel, and D.~Cremers, ``{LDSO: Direct Sparse Odometry
  with Loop Closure},'' in \emph{Proceedings of {IEEE} International Conference
  on Intelligent Robots and Systems (IROS)}, 2018, pp. 2198--2204.

\bibitem{Zubizarreta2020}
J.~Zubizarreta, I.~Aguinaga, and J.~M.~M. Montiel, ``{Direct Sparse Mapping},''
  \emph{{IEEE} Transactions on Robotics (T-RO)}, 2020.

\bibitem{Schonberger2016}
J.~L. Schonberger and J.~Frahm, ``{Structure-from-Motion Revisited},'' in
  \emph{Proceedings of the {IEEE} Conference on Computer Vision and Pattern
  Recognition (CVPR)}, 2016, pp. 4104--4113.

\bibitem{Sol}
J.~Sol, J.~Deray, and D.~Atchuthan, ``{A Micro Lie Theory for State Estimation
  in Robotics},'' \emph{arXiv:1812.01537v6}, 2018.

\end{thebibliography}
}

%
\begin{IEEEbiography}[{\includegraphics[width=1in,height=1.25in,clip,keepaspectratio]{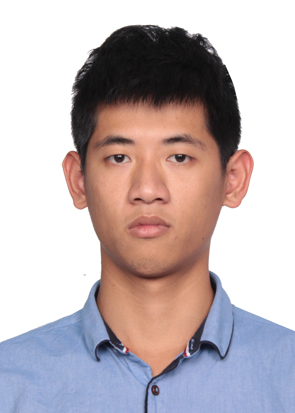}}]{Qiuyuan Wang}  received the B.E. degree in Internet of Things Engineering (IOT) from Beijing University of Technology, Beijing, China, in 2016, and the M.S. degree in the Key Laboratory of Machine Perception (Minister of Education) from Peking University, Beijing, China, in 2019. He is currently a researcher in Mobile SLAM group, Sensetime company, focusing on 3D vision, SLAM, and Deep Learning.
\end{IEEEbiography}

\begin{IEEEbiography}[{\includegraphics[width=1in,height=1.25in,clip,keepaspectratio]{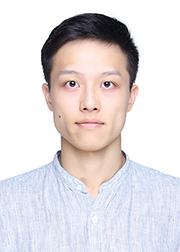}}]{Zike Yan} received the B.E. degree in measuring and control technology and instrument from the Harbin Institute of Technology, China, in 2014, and the M.S. degree in information and communication engineering from the Harbin Engineering University, China, in 2018. He is currently working toward the Ph.D. degree on the intersection of computer vision, robotics, and machine learning, focusing on scene representation and map-centric vision applications, with the Key Laboratory of Machine Perception (Minister of Education), Peking University.
\end{IEEEbiography}

\begin{IEEEbiography}[{\includegraphics[width=1in,height=1.25in,clip,keepaspectratio]{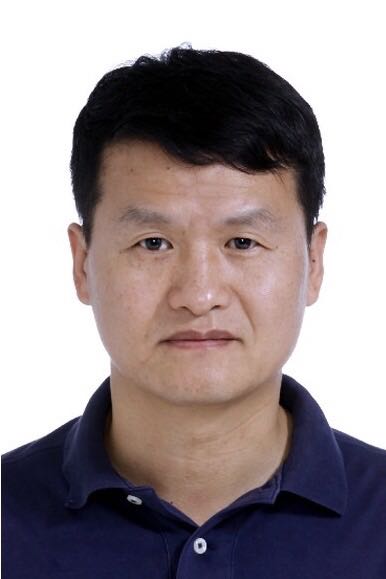}}]{Junqiu Wang}
	received the B.E. and M.E. degrees from Beijing Institute of Technology, Beijing, China, in 1992 and 1995, respectively, and the Ph.D degree from Peking University, Beijing, in 2006. He is currently with AVIC (Aviation Industry Corporation of China) Beijing Changcheng Aeronautical Measurement and Control Technology Research Institute, Beijing 10081, China. From 2006 to 2014, he was with the Institute of Scientific and Industrial Research, Osaka University, first as a post doc, then a specially appointed assistant professor, and a specially appointed associate professor. His current research interests are in image processing, computer vision, intelligent measurement, and robotics, including visual tracking, content-based image retrieval, image segmentation, vision-based localization, visual odometry, and SLAM. 
\end{IEEEbiography}

\begin{IEEEbiography}[{\includegraphics[width=1in,height=1.25in,clip,keepaspectratio]{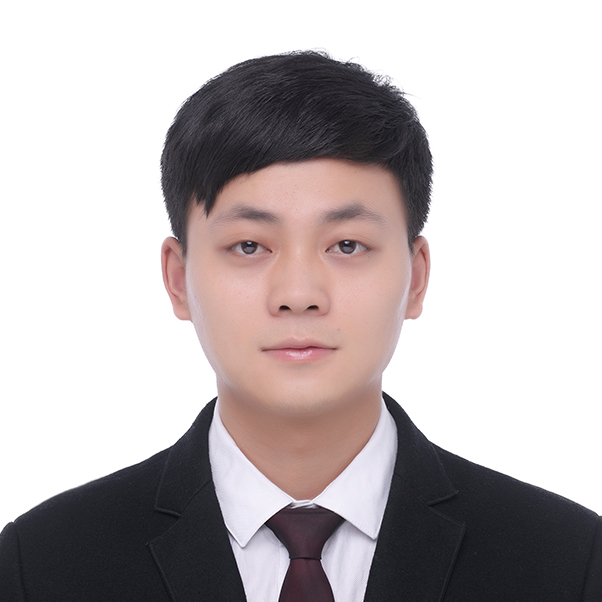}}] {Fei Xue}
	received his B.E. in 2016 from School of Electronics Engineering and Computer Science, Peking University, Beijing, China. In the same year, he joined the Key Laboratory of Machine Perception (Minister of Education) as a master student. His interests include visual odometry, simultaneous localization and mapping (SLAM), visual relocalization, and deep learning.
	
\end{IEEEbiography}

\begin{IEEEbiography}[{\includegraphics[width=1in,height=1.25in,clip,keepaspectratio]{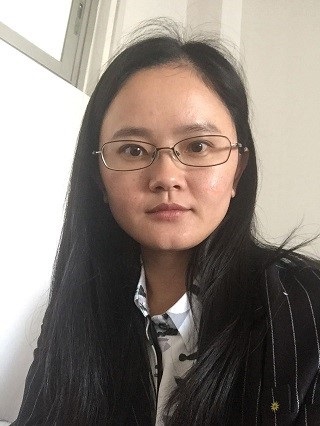}}]{Wei Ma}
	received her Ph.D. degree in Computer Science from Peking University, in 2009. She is currently an Associate Professor at Faculty of Information Technology, Beijing University of Technology. Her current research interests include image/video repairing, image/video semantic understanding, and 3D vision.
\end{IEEEbiography}

\begin{IEEEbiography}[{\includegraphics[width=1in,height=1.25in,clip,keepaspectratio]{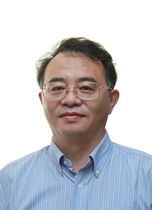}}]{Hongbin Zha}
	received the B.E. degree in electrical engineering from the Hefei University of Technology, China, in 1983 and the MS and PhD degrees in electrical engineering from Kyushu University, Japan, in 1987 and 1990, respectively. After working as a research associate at Kyushu Institute of Technology, he joined Kyushu University in 1991 as an associate professor. He was also a visiting professor in the Centre for Vision, Speech, and Signal Processing, Surrey University, Unite Kingdom, in 1999. Since 2000, he has been a professor at the Key Laboratory of Machine Perception (Ministry of Education), Peking University, Beijing, China. His research interests include computer vision, digital geometry processing, and robotics. He has published more than 300 technical publications in journals, books, and international conference proceedings.
\end{IEEEbiography}

%
%
%




\end{document}